\PassOptionsToPackage{table}{xcolor}
\documentclass{style}
\usepackage{marvosym}
\usepackage{amsmath,amssymb}
\usepackage{pifont}
\usepackage{array}
\usepackage{makecell}
\usepackage{textcomp}
\usepackage{tabularx}
\usepackage{enumitem}
\usepackage{xspace}
\usepackage{float}
\usepackage[bottom]{footmisc}
\usepackage{flafter}
\usepackage{listings}
\usepackage{url}
\usepackage{xurl}
\usepackage{ragged2e}
\usepackage{needspace}
\usepackage{fontawesome5}

\AtBeginDocument{%
}

\let\reportsection\section
\renewcommand{\section}{\Needspace*{5\baselineskip}\reportsection}

\definecolor{wedirectorblue}{HTML}{3F6FB5}
\definecolor{wedirectorgreen}{HTML}{3B7A57}
\definecolor{pendinggray}{HTML}{666666}
\definecolor{basegroup}{HTML}{EAF2FB}
\definecolor{editorgroup}{HTML}{F1ECFA}
\definecolor{traingroup}{HTML}{EAF6EE}
\definecolor{uppergroup}{HTML}{FFF1DE}
\definecolor{costgroup}{HTML}{F3F4F6}
\definecolor{codebg}{HTML}{F6F8FA}
\definecolor{codeframe}{HTML}{D0D7DE}
\definecolor{codekeyword}{HTML}{8250DF}
\definecolor{codestring}{HTML}{0A7A55}
\hypersetup{
  pdftitle={WeVisCut: The Whole Editing Room in One Agent -- From Narration to Long-Form Video},
  pdfauthor={Chenyu Mu, Wenming Weng, Qinrui Wang, Xingsong Ye, Hao Yu, Jun He, Chong Sun, Chen Li, Jing LYU, Erkun Yang, Cheng Deng}
}
\lstdefinestyle{supplementcode}{
  basicstyle=\ttfamily\scriptsize,
  backgroundcolor=\color{codebg},
  frame=single,
  rulecolor=\color{codeframe},
  framesep=5pt,
  xleftmargin=6pt,
  xrightmargin=6pt,
  breaklines=true,
  columns=fullflexible,
  keepspaces=true,
  showstringspaces=false,
  aboveskip=7pt,
  belowskip=7pt,
  captionpos=b
}

\definecolor{wedirectorblue}{RGB}{42,92,145}
\definecolor{tracebackground}{RGB}{247,249,252}
\definecolor{codebackground}{RGB}{241,244,248}
\definecolor{codeframe}{RGB}{205,213,223}

\newtcolorbox{promptbox}[2][]{%
  enhanced,
  breakable,
  colback=tracebackground,
  colframe=#2,
  coltitle=white,
  title={#1},
  fonttitle=\bfseries,
  boxrule=0.8pt,
  arc=1.5mm,
  left=2.5mm,
  right=2.5mm,
  top=2mm,
  bottom=2mm,
  before skip=8pt,
  after skip=8pt,
  before upper={%
    \RaggedRight
    \setlength{\parindent}{0pt}%
    \setlength{\parskip}{4pt}%
  },
}

\lstdefinestyle{toolcall}{
  basicstyle=\ttfamily\small,
  backgroundcolor=\color{codebackground},
  frame=single,
  rulecolor=\color{codeframe},
  framesep=4pt,
  xleftmargin=0pt,
  xrightmargin=0pt,
  aboveskip=4pt,
  belowskip=6pt,
  breaklines=true,
  breakatwhitespace=false,
  columns=fullflexible,
  keepspaces=true,
  showstringspaces=false,
  upquote=true,
  tabsize=2,
  postbreak=\mbox{\textcolor{gray}{$\hookrightarrow$}\space},
}

\newcommand{\tracestep}[2]{%
  \par
  \medskip
  \noindent\textbf{#1. #2}\par\nobreak
  \smallskip
}

\newcommand{\tracefield}[1]{%
  \par\noindent\textsc{#1.}\enspace
}

\newcommand{\cmark}{\textcolor{wedirectorgreen}{\ding{51}}}
\newcommand{\xmark}{\textcolor{red}{\ding{55}}}

\newcommand{\agent}{WeVisCut\xspace}

\newcommand{\rft}{\textsc{RFT}\xspace}

\newcommand{\gopd}{\textsc{Guided-OPD}\xspace}

\newcommand{\rcfdapo}{\textsc{RCF-DAPO}\xspace}
\newcommand{\hyperframes}{\textsc{HyperFrames}\xspace}
\newcommand{\tablefontsize}{\fontsize{9}{10.5}\selectfont}

\newcommand{\guidedopdcite}{\citep{li2026policy}}
\newcommand{\stepopdcite}{\citep{sun2026step}}
\title{WeVisCut: The Whole Editing Room in One Agent\\
From Narration to Long-Form Video}

\author[1,2,*]{Chenyu Mu}
\author[2,\dagger]{Wenming Weng}
\author[2,*]{Qinrui Wang}
\author[2]{Xingsong Ye}
\author[2]{Hao Yu}
\author[2]{Jun He}
\author[2,\ddagger]{Chong~Sun}
\author[2]{Chen Li}
\author[2]{Jing LYU}
\author[1,\ddagger]{Erkun Yang}
\author[1]{Cheng Deng}

\affiliation[1]{Xidian University}
\affiliation[2]{WeChat Vision, Tencent Inc.}
\newcommand\blfootnote[1]{%
  \begingroup
  \renewcommand\thefootnote{}%
  \begin{NoHyper}%
    \footnotetext{#1}%
  \end{NoHyper}%
  \endgroup
}
\newcommand{\ReportContributionNote}{%
  \textsuperscript{*}Work done as interns at WeChat.\enspace
  \textsuperscript{\ensuremath{\dagger}}Project leader.\enspace
  \textsuperscript{\ensuremath{\ddagger}}Corresponding authors.%
}

\abstract{

Producing long-form videos from narration requires interdependent decisions
that remain coherent across shots. Existing agentic editors often use
multi-stage workflows driven by general models, leaving the quality-cost
trade-off of compact, task-adapted policies unclear.
We address this gap with \textbf{\agent}, a long-form script-to-video system
that connects tool execution with policy learning. Its agent harness maintains
shared production state so that tool feedback informs subsequent decisions
and repairs.
Because finished videos do not expose these decisions, our four-stage data
pipeline combines editing inversion with verified replay to turn recovered
editing intent into executable supervision. Applied to 3,750 finished videos,
this pipeline yields over \textbf{130K} training examples.
We use this supervision to train a single 9B policy that coordinates
production and recovers from failures within the same harness.
Moreover, we introduce a 100-task benchmark and five metrics assessing editing
quality and delivery reliability.
Under a shared tool environment, \agent substantially outperforms its base
model and achieves comparable performance to Claude Opus 5 while using nearly
\textbf{33$\times$} fewer generated tokens per successful output minute.
It also outperforms task-matched native editors on our proposed metrics.

}

\checkdata[\raisebox{-0.15ex}{\includegraphics[width=1.05em,height=1.05em,keepaspectratio]{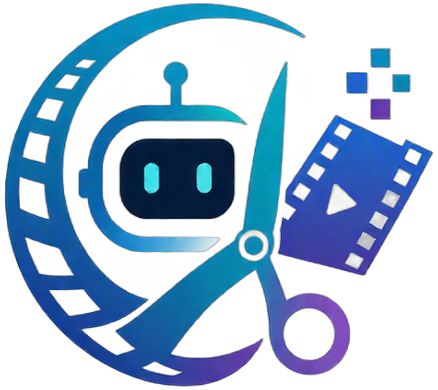}}~~Project Page]{%
  \href{https://tencent.github.io/WeVisCut/}{\texttt{https://tencent.github.io/WeVisCut}}}
\checkdata[\raisebox{-0.15ex}{\includegraphics[width=1.05em,height=1.05em,keepaspectratio]{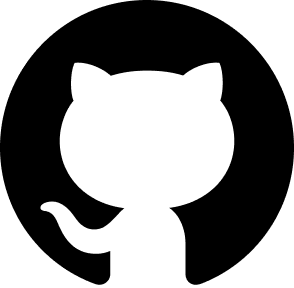}}~~Code]{%
  \href{https://github.com/Tencent/WeVisCut}{\texttt{https://github.com/Tencent/WeVisCut}}}
\checkdata[\raisebox{-0.15ex}{\includegraphics[width=1.05em,height=1.05em,keepaspectratio]{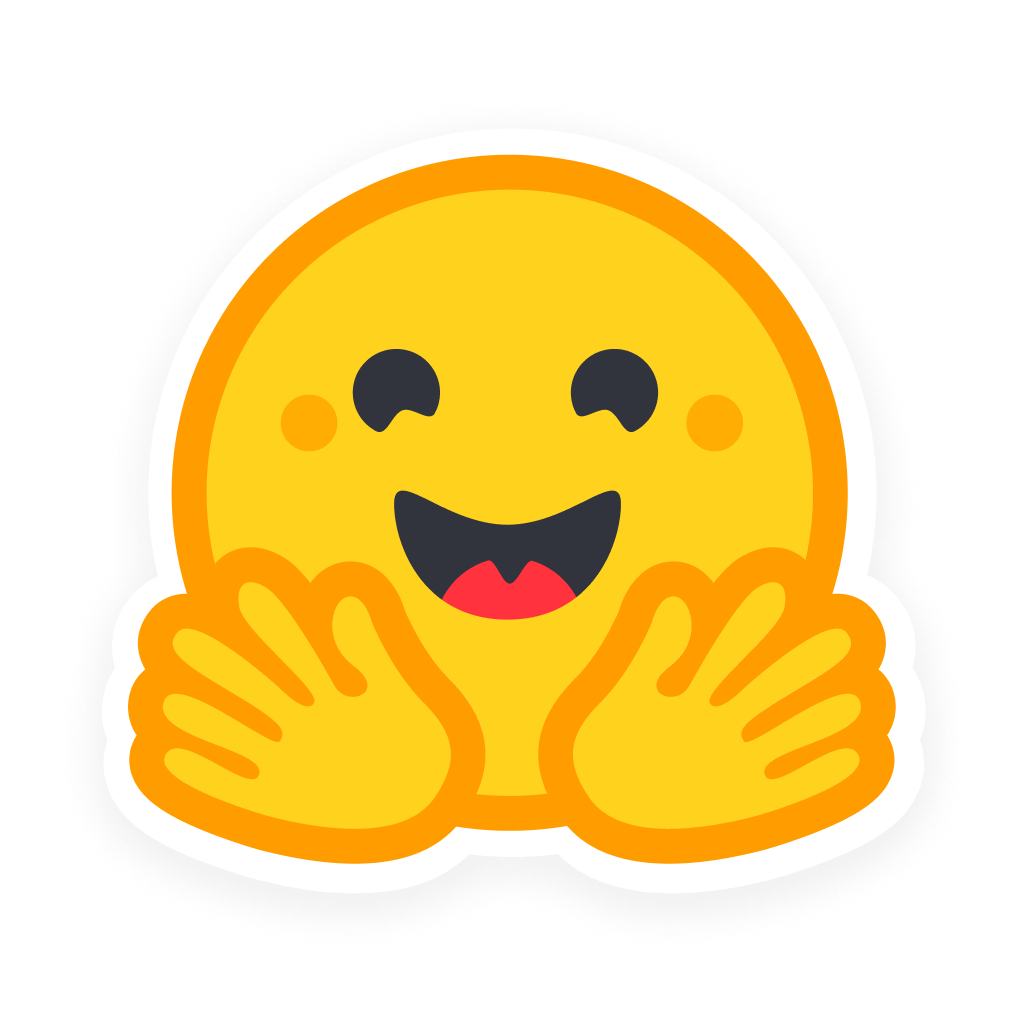}}~~Weights]{%
  \href{https://huggingface.co/tencent/WeVisCut}{\texttt{https://huggingface.co/tencent/WeVisCut}}}

\titleteaser{%
  \noindent\begin{minipage}{\linewidth}
    \centering
    \includegraphics[width=0.9\linewidth]{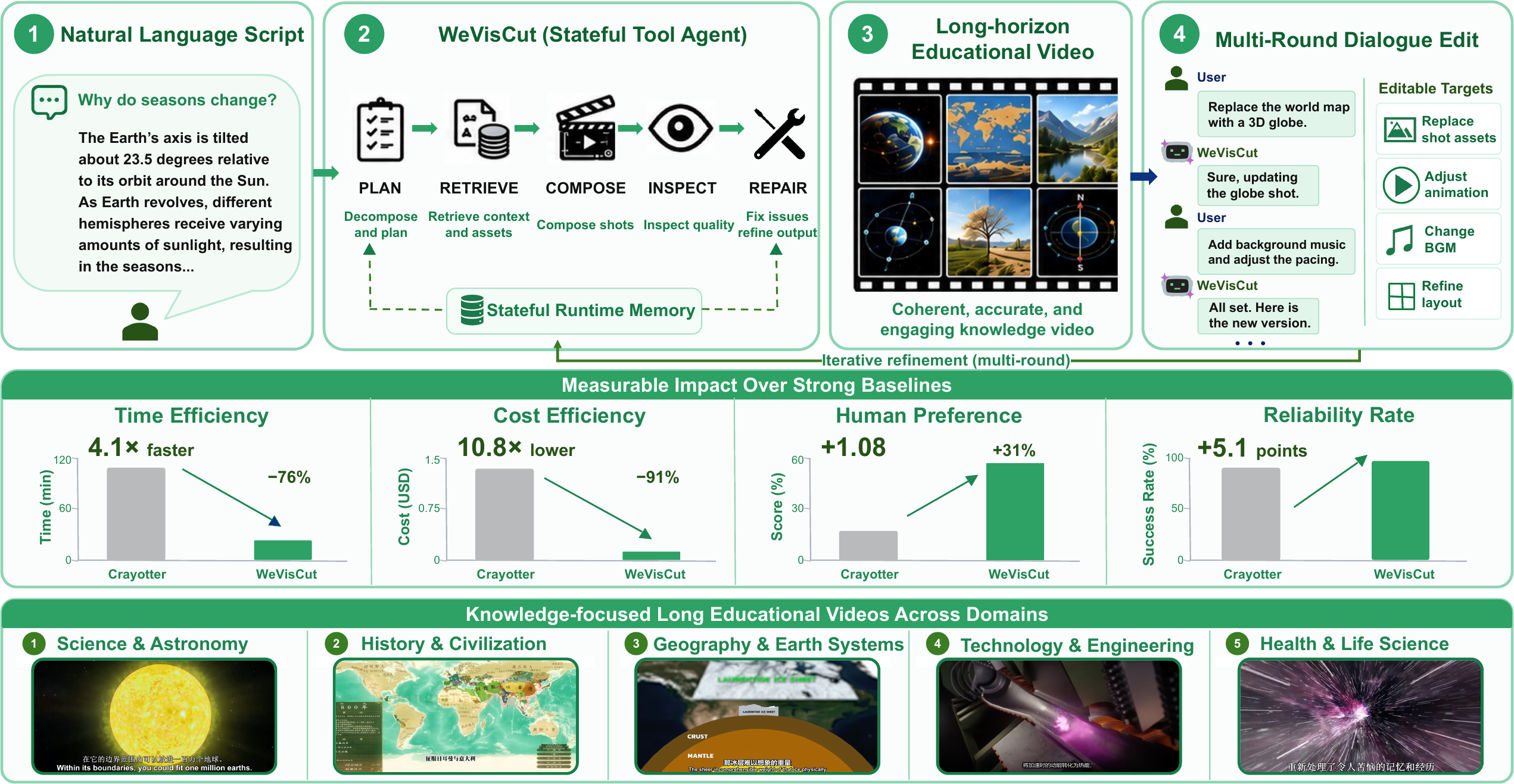}
    \captionsetup{hypcap=false,skip=2pt}
    \captionof{figure}{Overview of \textbf{WeVisCut}. 
    \textbf{Top}: Script-driven stateful video production (planning, retrieval, composition, inspection, repair) with multi-round edits. \textbf{Middle}: Baseline comparisons in time, cost, preference, and reliability. \textbf{Bottom}: Examples in science, history, geography, tech, and health.
    }
    \label{fig:introduction}
  \end{minipage}
}

\begin{document}
\enlargethispage{1mm}
\blfootnote{\ReportContributionNote}
\maketitle

\section{Introduction}

\begin{quote}
{\centering\normalfont\itshape
``The whole eloquence of cinema is achieved in the editing
room.''\par}
{\raggedleft \small --- Orson Welles\par}
\end{quote}

Creating a narrated long-form video involves turning a script into a
sequence of coherent shots. Planning, visual retrieval, narration synthesis,
and timeline assembly are interdependent. Changing a shot's duration affects
later timestamps, while unsuitable visual material may require both retrieval
and planning to be revisited across the affected shot sequence.

Recent agentic editors coordinate production through specialized agents and
tool workflows~\citep{zhao2026cutclaw,yan2026crayotter,zhang2026cineagents,
zhou2026videoagent,mu2026script,yu2026aurora}. Several systems use general
models with feedback or graph refinement. Yet orchestration alone does not
specify how to train a compact policy that maintains video quality at low
token cost. Crayotter trains a 9B editing model using final-video preferences
on manually constructed tasks~\citep{yan2026crayottergrpb}.
Such a policy needs a stateful environment for execution and recovery, together with
supervision grounded in that environment. Finished videos, however, omit the
tool calls, intermediate artifacts, and failures needed for such supervision.
The challenge is therefore
to connect stateful execution, executable supervision, and policy learning.

To address these challenges, we present \agent, which integrates a
script-to-video agent harness, executable data construction, and a task-adapted
policy. The harness (Figure~\ref{fig:framework}) maintains shared production
state and validates tool actions. It exposes execution feedback for inspection,
bounded repair, and replanning while preserving unaffected artifacts
from earlier successful production steps.

Within this harness, we construct executable supervision through the
four-stage pipeline in Figure~\ref{fig:data}. We first recover script--shot
intent through editing inversion and execute it through verified replay.
When replay reveals defects, we apply observe--reflect--repair for bounded
corrections that preserve successful artifacts. We then use trajectory
materialization to organize the recorded executions into training examples
with tool feedback and repair outcomes. From 3,750 video projects, this
process yields over 130K examples.

Using these examples, we train a single 9B policy across production roles in
three phases (Figure~\ref{fig:overview}). In Phase~1, we establish tool use
and recovery through unified supervised fine-tuning~\citep{zeng2023agenttuning}
and runtime-verified rejection-sampled fine-tuning~\citep{yuan2023scaling}.
In Phase~2, we apply Guided-OPD~\guidedopdcite{} to learn recovery on
policy-reached states while phasing out teacher intervention.
In Phase~3, we use \rcfdapo, our role- and context-factored
DAPO~\citep{yu2025dapo}, to optimize decisions by comparing runtime outcomes
within matched roles and contexts.

Moreover, we introduce a benchmark of 100 script-to-video tasks and five complementary
metrics to evaluate editing quality and delivery reliability. These metrics
relate rendered videos to their scripts and production plans, so evaluation
goes beyond successful rendering alone. In a shared tool environment, our
policy improves over its base model by \textbf{16.0} points in Overall Score
and achieves comparable performance to Claude Opus 5 while using nearly
\textbf{33$\times$} fewer generated tokens per successful output minute.
Our complete system also outperforms task-matched native editors.

We summarize our contributions as follows.
\begin{enumerate}[leftmargin=12pt,itemsep=1pt,topsep=2pt]
    \item We design a script-to-video agent harness that maintains shared
    production state, validates tool actions, and supports inspection,
    bounded repair, and final delivery.

    \item We build a four-stage data pipeline based on editing inversion and
    verified replay, converting finished videos into executable training examples.

    \item We train a unified 9B policy for tool use and recovery through
    sequential SFT, runtime-verified RFT, Guided-OPD, and \rcfdapo{}
    within a shared production harness.

    \item We introduce a 100-task benchmark and five metrics. \agent improves
    its base policy, outperforms task-matched systems, and achieves comparable
    quality to Claude Opus 5 at markedly lower cost.
\end{enumerate}

\section{Related Work}

\textbf{Tool agents and long-horizon post-training.}
Producing long-form videos requires agents to choose tools and revise plans
from execution feedback. Earlier methods range from interleaved reasoning and
action~\citep{yao2023react} to learned API calls
~\citep{schick2023toolformer,qin2023toolllm,patil2023gorilla} and runnable
tool dialogues~\citep{li2023apibank}. To
train such behavior, researchers curate interactions and synthesize
instructions or tool calls~\citep{zeng2023agenttuning,chen2024agentflan,
mitra2024agentinstruct,liu2024toolace}. Reflection and reusable skills also
support recovery~\citep{shinn2023reflexion,wang2023voyager}. Closer to our
data pipeline, AgentTrek builds verified web trajectories from
tutorials~\citep{xu2024agenttrek}. Finished videos, however, do not record
editing actions. We therefore infer script--shot intent and replay it with
production tools to obtain executable trajectories. Long-horizon training
also needs to handle policy-generated states. DAgger~\citep{ross2011dagger} collects expert labels for
learner states, while generalized
knowledge distillation trains on student-generated sequences
~\citep{agarwal2023gkd}. Guided-OPD mixes teacher and student
turns, while TurnOPD and Step-Level OPD refine supervision at turn and step
boundaries~\citep{li2026policy,zhou2026turnopd,sun2026step}. We apply
Guided-OPD with truncated teacher logits. For reward optimization, GRPO and
DAPO build on PPO-style updates with group-relative comparisons
~\citep{schulman2017ppo,shao2024deepseekmath,yu2025dapo}. We adapt DAPO for
our multi-role policy as \rcfdapo, comparing
rewards only within matched roles and decision contexts.

\textbf{Video editing.}
Video editing systems range from user-assisted tools to autonomous agents.
LAVE keeps users in the editing loop~\citep{wang2024lave}. In more automated
settings, CutClaw and CineAgents handle music-synchronized editing and
cinematic compilation~\citep{zhao2026cutclaw,zhang2026cineagents}.
The original Crayotter uses a traceable multi-agent workflow, while VideoAgent
assembles specialists dynamically~\citep{yan2026crayotter,zhou2026videoagent}. Beyond
workflow design, task-specific training has also been explored. ``The Script
is All You Need'' trains a ScripterAgent but uses a separate DirectorAgent
~\citep{mu2026script}. Aurora trains a planner for local edits
~\citep{yu2026aurora}. Building on the Crayotter environment, subsequent work
trains a 9B policy with final-video preferences on manually constructed
tasks~\citep{yan2026crayottergrpb}. In contrast, \agent learns one role-routed
policy for fixed-narration production from executable trajectories built
through editing inversion, verified replay, observe--reflect--repair, and
trajectory materialization.

\textbf{Agent and video evaluation.}
Evaluation of narration-driven video production needs to consider both the
rendered video and the production process. VBench, VideoScore, and VMBench
assess generated-video quality and motion
~\citep{huang2024vbench,he2024videoscore,ling2025vmbench}, while Goku-Bench
tests instruction-based edits~\citep{liang2026goku}. Script--video timing has
also been evaluated~\citep{mu2026script}. Complementary agent evaluation
ranges from recorded web and mobile trajectories
~\citep{deng2023mind2web,rawles2023android} to interactive shopping, browser,
and desktop environments that validate action outcomes
~\citep{yao2022webshop,zhou2023webarena,koh2024visualwebarena,
drouin2024workarena,xie2024osworld}. Software-repair benchmarks use repository
tests~\citep{jimenez2024swebench}, while multi-turn and tool-oriented suites
measure progress, valid transitions, and task constraints
~\citep{liu2023agentbench,ma2024agentboard,lu2024toolsandbox,yao2024taubench}.
We combine these perspectives in a benchmark of 100 script-to-video tasks,
linking scripts, shot plans, tool trajectories, and rendered videos. Five
metrics are introduced to assess script fidelity, visual--narration alignment, timeline plan
realization, long-horizon consistency, and end-to-end delivery success.

\section{\agent Agent Framework}
\label{sec:agent-framework}

Figure~\ref{fig:framework} presents the \agent agent framework for long-form
script-to-video production. A shared policy makes editing decisions, while a
stateful runtime executes tools and maintains the resulting production
artifacts. Their interaction connects the narration script to a rendered video
through an inspectable process that supports revision when execution reveals
problems. We first define the task and policy--runtime interface, then describe
the harness that implements this process.

\subsection{Script-to-Video Task}

A task input $x$ consists of a fixed narration script $S$, target aspect ratio
$a^\star$, feasible duration interval $D^\star$, optional source assets, and a
versioned shared tool space $\mathcal T$. The goal is to render a video $V$
satisfying the delivery constraints without rewriting the narration. The
outputs also include a typed shot plan $P$, specifying narration coverage,
visual content, and timing, and an execution trajectory $\tau$ recording calls
and returns throughout the production process (Appendix~\ref{app:experiments}).

Production uses tools that accept typed arguments and return artifacts or
execution feedback. At call $t$, a single policy $\pi_\theta$ with shared
parameters $\theta$ serves a role $r_t\in\{\mathrm O,\mathrm G,\mathrm C\}$.
The orchestrator ($\mathrm O$) selects tools and arguments, the generator
($\mathrm G$) supplies structured content, and the critic ($\mathrm C$)
inspects artifacts and requests or verifies repairs based on recorded execution feedback.

\begin{figure}[htbp]
    \centering
    \includegraphics[width=\linewidth]{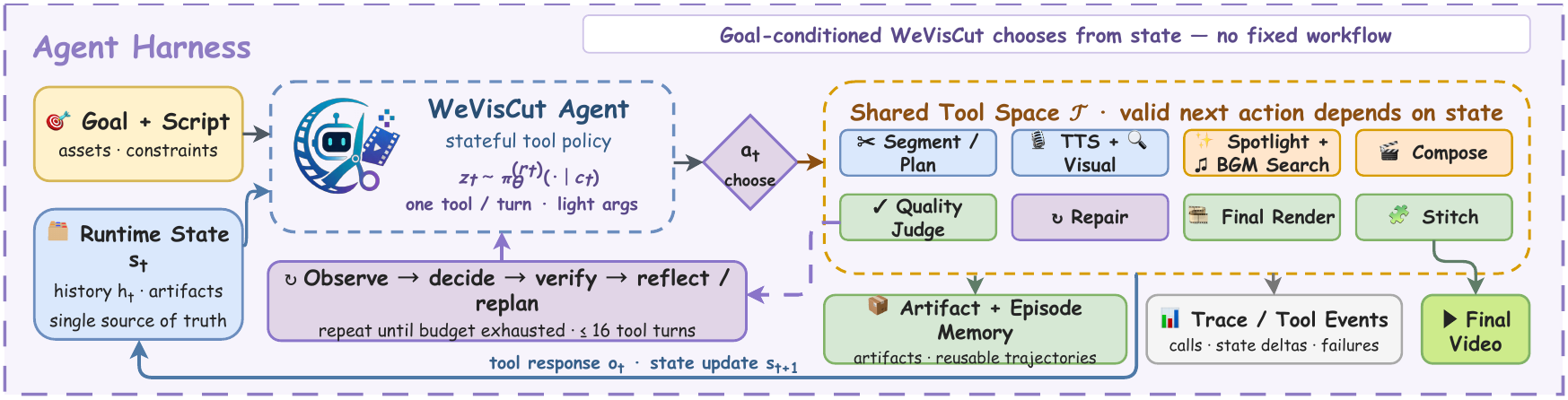}
    \caption{The WeVisCut agent harness with stateful tool selection. A shared policy chooses legal
    tools from the evolving production state. Validated returns update
    artifacts and inform inspection, local repair, and replanning before
    final stitching and delivery.}
    \label{fig:framework}
\end{figure}

At each call, the role context $c_t$ combines the task input, role, a view of
the production state $s_t$ (shot plan, artifacts, and progress), relevant
call-and-return history $h_t$, and tool schemas. The policy produces a complete
role output $z_t$, which the runtime
$\mathcal E$ processes under a tool contract $u_t$:
\begin{equation}
z_t\sim\pi_\theta^{(r_t)}(\cdot\mid c_t),\qquad
(o_t,s_{t+1})\sim\mathcal E_{r_t,u_t}(\cdot\mid s_t,z_t).
\label{eq:process}
\end{equation}
The contract specifies the tool, required fields, and value types. The runtime
returns observation $o_t$, reporting results or blockers, and the updated
production state $s_{t+1}$. For orchestrator calls, the role output $z_t$
specifies a tool action $a_t$ containing the selected tool and arguments
(Figure~\ref{fig:framework}).

\subsection{Harness Design}

The harness in Figure~\ref{fig:framework} implements the interaction in
Equation~\ref{eq:process} through a shared tool space, persistent production
state, and runtime checks. Each orchestrator turn selects one tool, which may
process multiple shots and invoke generator or critic calls. Tool selection
depends on the current production state, including available artifacts and blockers,
rather than a predetermined sequence.

\textbf{Production flow and tool system.}
Starting from the narration script $S$, segmentation produces production
segments, and shot planning expands them into the shot plan $P$. Shot records
connect narration spans to visual queries, target durations, and spotlight
assignments. Visual retrieval and text-to-speech (TTS) tools use these records
to obtain shot assets and narration audio. For spotlight shots, decoration
tools select verified \hyperframes templates~\citep{heygen2026hyperframes} and
fill typed slots, allowing the policy to control presentation through
structured content rather than raw layout code. Music search supplies
background music (BGM) for the final mix. Timeline composition combines shot
assets, narration audio, and decorations, while final rendering produces the
shot videos required for stitching. Quality judging and repair tools allow
the policy to revisit these intermediate products before delivery. This
progression describes artifact dependencies, not an obligatory ordering of
every tool call.

\textbf{Persistent state and selective context.}
These tools communicate through the production state $s_t$. Global records
track narration coverage, shot order, and timing, while per-shot records
associate assets, audio, templates, renders, and unresolved issues with shot
IDs. Planned content remains distinct from artifacts that have actually been
produced. Episode memory retains these artifacts, and the event trace
preserves calls, returns, and failures. The initial observation exposes the
full production state. Subsequent role contexts $c_t$ contain the latest
runtime observation, state changes, a progress summary, and relevant local
records. Separating persistent storage from model context avoids repeatedly
serializing the entire project while preserving the information needed for
later inspection and repair.

\textbf{Validated execution and actionable feedback.}
Given a tool action $a_t$, the runtime identifies the tool contract $u_t$ and
checks argument schemas, artifact references, and execution prerequisites.
When structured content or inspection is required, it supplies this contract
in the generator's or critic's role context. Only the runtime commits
validated results to the production state, whereas
rejected calls return typed blockers that identify what must be corrected or
produced next. For example, a stitching request cannot substitute for missing
shot renders or background music. The runtime reports these missing
prerequisites, allowing the orchestrator to select the necessary production
tools before retrying. This separation keeps model decisions grounded in
execution evidence rather than treating a proposed plan as completed work.

\textbf{Artifact-grounded inspection and bounded repair.}
Execution feedback also determines when revision is needed. Quality inspection
combines structured checks of shot records, timing, and render status with a
VLM-Judge over spotlight renders. The critic identifies affected shots and the
defects that a repair should address. Repair tools then revise the relevant
content and recheck affected downstream artifacts while retaining still-valid
results. For example, correcting a decoration need not regenerate unchanged
narration audio. Repair limits bound repeated correction attempts, but
exhausting this budget does not remove the artifact prerequisites for delivery.
Quality warnings and missing production artifacts therefore remain distinct
conditions. The representative trajectory in Appendix~\ref{app:tool-flow}
illustrates how inspection, blocked calls, and targeted follow-up actions
interact within this loop.

\textbf{Verified delivery and trace retention.}
When the required artifacts are available, the policy requests final
stitching. Reaching the call limit also triggers a delivery attempt from the
available production state. Before stitching, the runtime checks the required
shot renders, narration audio, and background music. After stitching, it
validates the video and delivery manifest. Missing prerequisites are returned
as blockers while further calls are available. A run that terminates without a
validated final video is recorded as incomplete, with its execution trace
retained for diagnosis. The resulting production records support executable
data construction in Section~\ref{sec:data-construction} and policy training in
Section~\ref{sec:training}. This production harness is separate from the
read-only evaluation harness in Appendix~\ref{app:harness}.

\section{Data Construction}
\label{sec:data-construction}

The harness records execution, but finished videos expose only narration
and shot structure, not the tool calls, feedback, and repairs that produced
them. To train the shared policy, we recover editing intent and replay it
under the same state and action contract, constructing role-specific
examples grounded in real tool returns from executing the recovered intent.

Our source data consists of finished narration-driven videos collected from
Bilibili, spanning technology and digital products, business and finance,
natural science, and social sciences and humanities. The admitted set contains
3,750 video projects.

\subsection{Executable Trajectory Construction}

\begin{figure}[!ht]
    \centering
    \includegraphics[width=\linewidth]{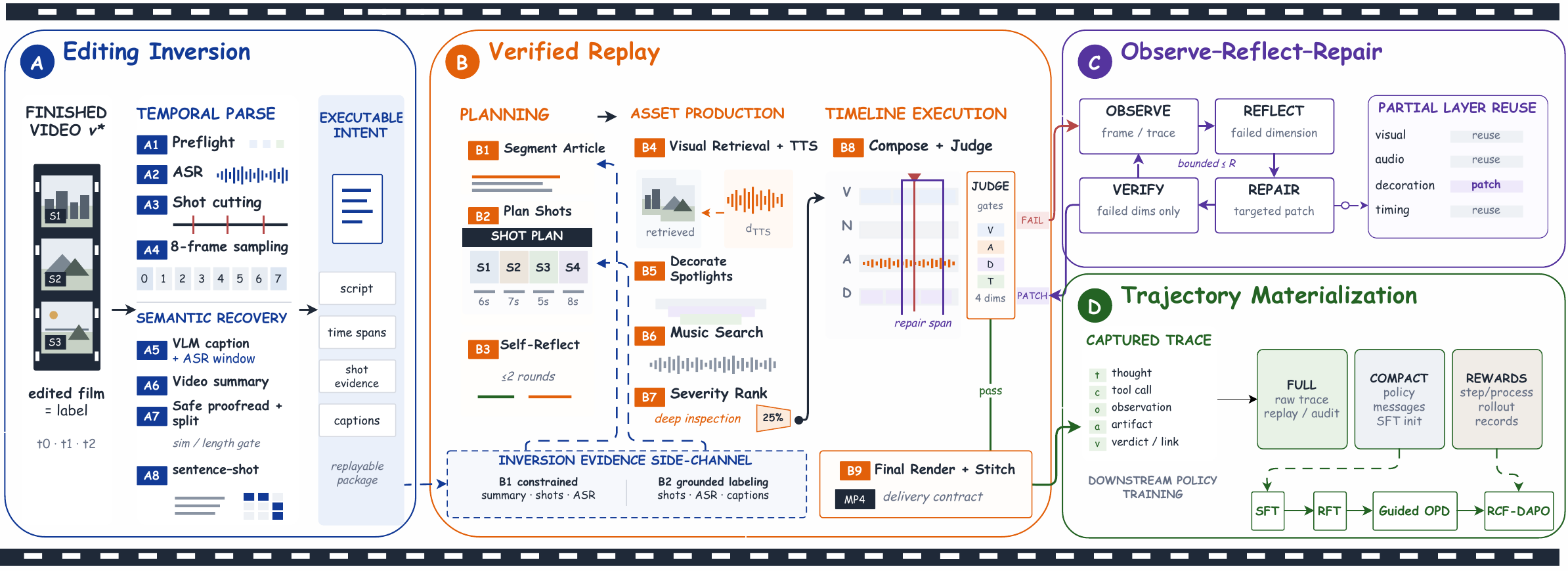}

    \caption{Executable data construction through editing
    inversion, verified replay, observe--reflect--repair, and trajectory
    materialization. (A) Editing inversion recovers aligned narration and shot
    evidence from a finished video. (B) Verified replay executes this intent
    with deployment-aligned production tools. (C) Observe--reflect--repair
    performs bounded inspection and targeted repair to preserve successful
    artifacts. (D) Trajectory materialization organizes recorded tool
    interactions, feedback, and repair links into role-specific training
    examples and rollout records.}
    \label{fig:data}
\end{figure}

To convert these videos into executable supervision, we use a four-stage
pipeline comprising editing inversion, verified replay, observe--reflect--repair,
and trajectory materialization (Figure~\ref{fig:data}).

\textbf{Editing Inversion.}
We first recover script--shot intent from the source videos
(Figure~\ref{fig:data}A), including timestamped narration, shot structure, and
complete sentence--shot alignment for every admitted project. Narration
provides temporal anchors, while aligned shots indicate how the content is
organized visually. Together, this evidence forms a structured storyboard
describing what each shot should communicate, how long it should last, and
which production constraints it should satisfy. This process recovers the
intended video structure, not the original editor's tool actions.

\textbf{Verified Replay.}
The recovered storyboard specifies what to produce, but does not provide
execution feedback. Verified replay (Figure~\ref{fig:data}B) therefore executes
this intent through a production toolchain aligned with the deployment state
and action contract.
Schema-checked tools maintain a shared production state across planning
(article segmentation, shot planning, and self-reflection), asset production
(visual retrieval with TTS, spotlight decoration, music search, and severity
ranking), and timeline execution (composition with judging, final rendering,
and stitching). Each real tool
return updates this state before the next decision. The runtime verifies
asset references, audio durations, template inputs, timeline positions, and
rendered outputs against the artifacts actually produced. Spotlight shots use
verified \hyperframes templates~\citep{heygen2026hyperframes} that separate
layout, motion, and visual style. The policy selects a template and fills
typed slots instead of generating raw HTML, CSS, or JavaScript.

\textbf{Observe--Reflect--Repair.}
During replay, quality inspection combines structured shot, timing,
decoration-slot, and render-status checks with a VLM-Judge over spotlight
renders. Using these inspection results, we locate production problems,
apply bounded local repairs supported by the toolchain, and verify these
changes while preserving successful artifacts (Figure~\ref{fig:data}C).

\textbf{Trajectory Materialization.}
These executions produce trajectories that retain pre-action states, tool
names and arguments, real returns, created artifacts, quality verdicts, and
failure--repair links. Missing arguments, unresolved references, invalid stage
order, and failed delivery checks remain visible rather than being rewritten
as successes. We then organize these trajectories into role-specific examples
and rollout records (Figure~\ref{fig:data}D).

\subsection{Training Examples and Data Split}

These examples provide role-specific supervision under the framework in
Figure~\ref{fig:framework}. Orchestrator samples supervise tool selection and
structured arguments, while generator samples cover script segments, shot
plans, retrieval queries, and template content. Critic samples provide failure
localization, repair requests, and verification decisions. Runtime
observations remain context, with loss applied only to model-authored tokens.

We partition by source project, keeping all its examples and repair
records together. Held-out projects supply the benchmark in
Section~\ref{sec:experimental-setup} and are excluded from training. The
134,771 examples supervise a single Qwen3.5-9B policy with shared production
state and tool vocabulary across all three roles.

\section{Training \agent}
\label{sec:training}

\begin{figure}[!t]
    \centering
    \includegraphics[width=\linewidth]{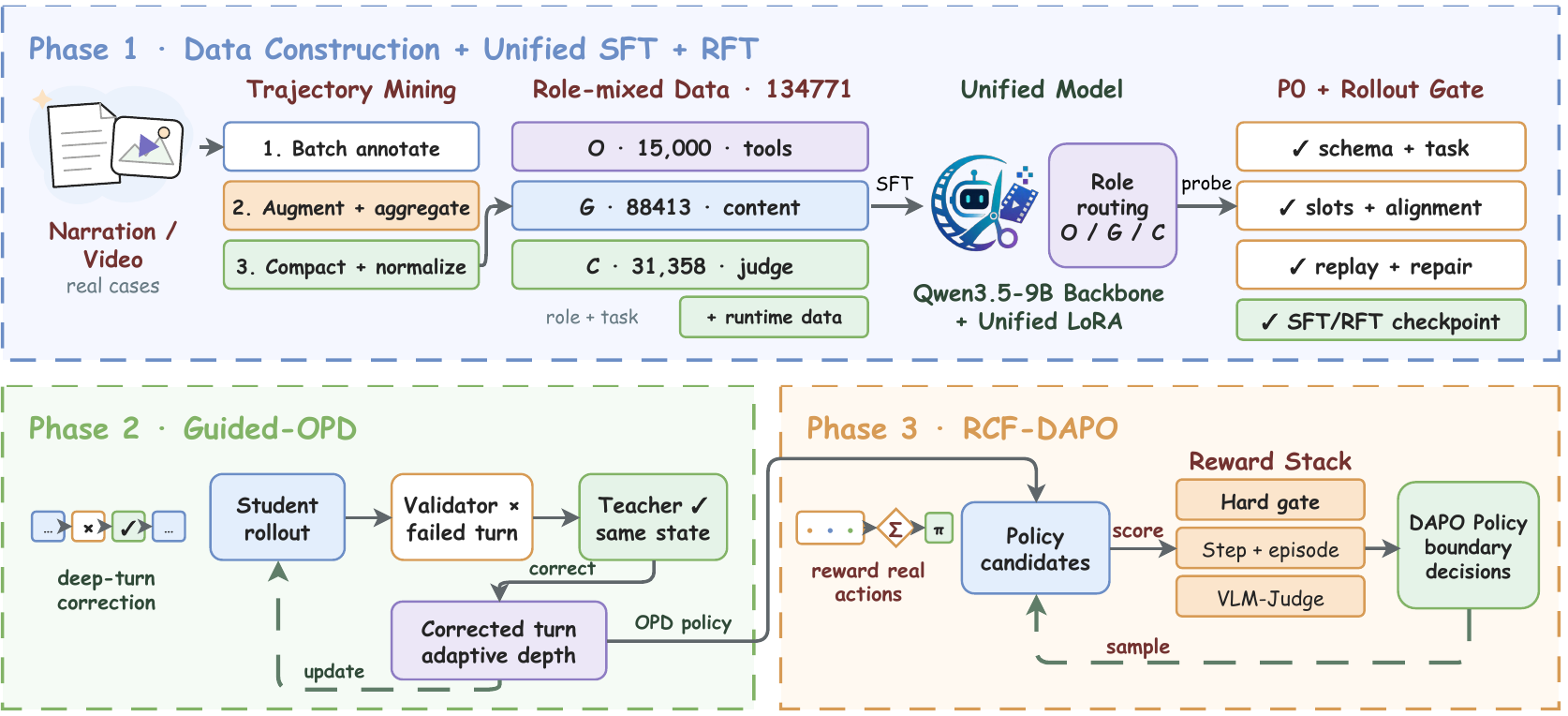}
    \caption{Three training phases of \agent. Phase 1 learns tool use and
    recovery through unified SFT and runtime-verified RFT. Phase 2 applies
    Guided-OPD with decreasing teacher intervention. Phase 3 uses
    \rcfdapo to compare alternatives within matched roles and decision
    contexts.}
    \label{fig:overview}
\end{figure}

The executable examples provide initial supervision, but \agent also needs
to handle the states produced by its own decisions. Figure~\ref{fig:overview}
summarizes three training phases, comprising unified SFT and runtime-verified
RFT, recovery learning with Guided-OPD, and decision optimization with
\rcfdapo. The policy interaction and agent harness are defined in
Section~\ref{sec:agent-framework}.

\subsection{Phase 1: Unified SFT and RFT}

We first train Qwen3.5-9B with LoRA~\citep{hu2021lora} on the 134,771
examples constructed above. Following AgentTuning and
Agent-FLAN~\citep{zeng2023agenttuning,chen2024agentflan}, we supervise model
responses while retaining tool feedback as input.

Let $\mathcal D_{\rm SFT}$ contain $N_{\rm SFT}$ training examples.
We index examples by $e$ and tokens within an example by $j$.
The conversation $y_e=(y_{e,1},\ldots,y_{e,L_e})$ has $L_e$ tokens,
and $y_{e,<j}$ is the prefix before position $j$. Before each response,
this prefix encodes the role context $c_t$, including instructions and
available tool feedback. The loss mask $m_{e,j}$ is one for tokens in the
target model response and zero for instructions, tool-return messages, and
padding tokens added to equalize sequence lengths for batching. Thus only
response tokens contribute to the supervised fine-tuning loss
\begin{equation}
\mathcal L_{\rm SFT}
=-\frac{\displaystyle\sum_{e=1}^{N_{\rm SFT}}\sum_{j=1}^{L_e}
m_{e,j}\log\pi_\theta(y_{e,j}\mid y_{e,<j})}
{\displaystyle\sum_{e=1}^{N_{\rm SFT}}\sum_{j=1}^{L_e}m_{e,j}}.
\label{eq:sft}
\end{equation}
The denominator counts the supervised tokens, so the loss averages their
negative log probabilities across all supervised model responses in the corpus.

SFT teaches the recorded production steps, but the student can encounter
new failures when executing tasks on its own. We collect additional
supervision through runtime-verified rejection-sampled fine-tuning (RFT).
At the first recoverable failure, we restore the last replayable state before
the error and keep the valid earlier steps. The teacher regenerates the
complete remaining sequence from that state. Both student-only successes and
teacher-assisted repairs are replayed and admitted only if they pass the
runtime checks and a quality threshold selected on development tasks.
We track and balance these two sources of supervision. Failures that cannot
be restored enter a bounded pool for later rollout sampling.
Appendix~\ref{app:opd} gives the admission rules for both sources of supervision.

\subsection{Phase 2: Recovery Learning with Guided-OPD}

We next apply Guided-OPD~\guidedopdcite to learn decisions while interacting
with the production environment. At training update $n$, the teacher
generates a complete role output with probability $\beta_n$, and the
student generates it with probability $1-\beta_n$. At call $t$, we draw
$q_t\sim\operatorname{Bernoulli}(\beta_n)$ and use
\begin{equation}
z_t\sim
\begin{cases}
\pi_\phi^{(r_t)}(\cdot\mid c_t),&q_t=1,\\
\pi_\theta^{(r_t)}(\cdot\mid c_t),&q_t=0
\end{cases}.
\label{eq:opd}
\end{equation}
Here $\phi$ is the frozen teacher's parameter set. Both policies receive
the same role context $c_t$. As in Equation~\ref{eq:process}, the complete
output $z_t$ is executed before the next call, when control is sampled again.

Both branches train the student. Teacher outputs use forward KL against
the teacher distribution renormalized over its $K_{\rm vocab}$ most likely
tokens, reducing the cost of distillation. Student outputs use reverse KL
on their own generated prefixes. We gradually reduce $\beta_n$ from a
value below one to zero, so training ends with fully student-controlled
executions. Appendix~\ref{app:opd} specifies the token masks, KL losses,
and intervention schedule used throughout this training phase.

\subsection{Phase 3: Decision Optimization with \rcfdapo}

Finally, we compare alternative outputs for the same production decision.
Our Role- and Context-Factored DAPO (\rcfdapo) adapts
DAPO~\citep{yu2025dapo} by restricting reward comparisons to the same role
and production context. At a fixed role call, we sample candidate outputs
$z_i$, where $i$ indexes alternatives in the comparison group. For example,
generator candidates may be different retrieval queries for the same shot.
For orchestrator calls, we restore the
same production state, sample alternative next actions, and execute each
with a frozen continuation policy, the same tool registry, budget, and
paired random seeds. Generator and critic outputs are evaluated under the
same local request, artifact, and tool contract.

Each candidate receives a role-specific reward $R_i^{(r)}$.
Orchestrator rewards combine successful delivery and final quality with
partial credit for persistent progress on failed executions. Generator
rewards assess schema validity, requirement coverage, and content usability.
Critic rewards assess diagnosis, repair utility, and verification, with a
penalty for unnecessary repairs. The scoring rules, rubrics, and weights
are fixed. Appendix~\ref{app:dapo} defines the complete reward formulas.

Within a group with matching conditions $\kappa$, we convert rewards into
relative advantages
\begin{equation}
A_i^{(r)}=\operatorname{sg}\!\left[
\frac{R_i^{(r)}-\mu_\kappa^{(r)}}{\sigma_\kappa^{(r)}+\epsilon}\right],
\label{eq:dapo-advantage}
\end{equation}
where $\mu_\kappa^{(r)}$ and $\sigma_\kappa^{(r)}$ are the group's mean
reward and standard deviation, $\epsilon>0$ prevents division by zero,
and $\operatorname{sg}$ stops gradients through the advantages.
Above-mean rewards yield positive advantages, and below-mean rewards yield
negative ones. Groups with insufficient
reward variation are resampled, with quotas and a replay queue maintaining
coverage. Contexts are collected from the current policy with gradients
stopped, and role losses are balanced after token-level normalization.
We retain DAPO's asymmetric Clip-Higher and soft overlong shaping.
Appendix~\ref{app:dapo} details the sampling and updates.

\section{Experiments}

We evaluate whether task-specific training improves script-to-video production
and whether these gains extend to complete editing systems. We first compare
policies in a shared tool environment, then compare \agent with editors using
their native workflows. Separating these comparisons distinguishes policy
performance from differences in tools and orchestration. We further examine
the cumulative training stages and assess the rendered videos through blinded
human evaluation.

\subsection{Experimental Setup}
\label{sec:experimental-setup}

\textbf{Benchmark and implementation.}
We construct a fixed benchmark of 100 script-to-video tasks from the held-out
source projects in Section~\ref{sec:data-construction}. Each task specifies a
narration script, source pool, target duration, and aspect ratio. All systems
receive identical task inputs and delivery constraints. Our
trainable variants share the Qwen3.5-9B backbone, training split, decoding
configuration, tool registry, and rendering budget. The Guided-OPD and
\rcfdapo training configurations are given in Appendix
Table~\ref{tab:training-config}.

\textbf{Comparison methods.}
For the policy comparison, all models use the same instructions, tool schemas,
runtime, and rendering budget. We evaluate Qwen3.5-9B, Qwen3.6-27B, and
Qwen3.6-35B-A3B~\citep{qwen3.5,qwen3.6-27b,qwen3.6-35b-a3b}, together with
Qwen3.5-397B-A17B, Gemini 3.1 Pro~\citep{gemini31pro}, and a fixed
Claude Opus 5 snapshot. These reference models receive no task-specific
post-training. Our variants differ only in the training stages applied to the
shared 9B backbone within the same production environment.

We compare \agent with Crayotter~\citep{yan2026crayottergrpb} and the
commentary-capable VideoAgent workflow~\citep{zhou2026videoagent}, each using
its released tools and orchestration. Adapters may translate file formats and
export evaluation artifacts, but may not retrieve assets, write narration,
create plans, or repair outputs. CutClaw, CineAgents, and
DirectorAgent~\citep{zhao2026cutclaw,zhang2026cineagents,mu2026script} are
excluded from this quantitative comparison because their native workflows
omit or replace parts of the fixed-narration, fixed-source task contract.
Appendix Table~\ref{tab:baseline-contract} shows whether each system
supports the required inputs, shot plans, and execution.

\textbf{Evaluation metrics.}
We propose five complementary metrics for script-to-video production that
assess the relationship between the source script, editing plan, execution
trace, and rendered video. All scores lie in $[0,100]$, with higher values
indicating better performance. Script Content Fidelity (SCF) measures
the coverage, faithfulness, and ordering of narration claims relative to the
source script. Visual--Narration Semantic Match (VNSM) assesses whether
rendered visuals support the concurrent narration. Timeline Plan Realization
(TPR) compares planned and realized shots in presence, duration, and order.
Long-Horizon Consistency (LHC) measures late-stage shot-health retention and
asset diversity. End-to-End Rendering Yield (ERY) is the fraction of held-out
tasks producing a decodable, sufficiently complete video with the required
non-silent narration. Overall Score is the equally weighted mean of the five
reported metric means, used only as a descriptive summary
for comparing aggregate task performance.

Production Cost Efficiency (PCE) is reported separately as tool calls and
generated tokens per successful output minute. For native systems, we instead
report wall-clock minutes and estimated USD per successful output minute,
since the underlying tools and services differ. Both measures include failed
attempts in the cost numerator, and lower values are better. Definitions,
eligibility rules, and missing-data treatment are given in
Appendix~\ref{app:metrics}.

\textbf{Evaluation protocol.}
Each system is evaluated in five independent runs over the 100 tasks. We
report across-run means and standard deviations for the five component
metrics, with no uncertainty estimate for Overall Score. Permissions and
retry limits are fixed within the shared environment, and thresholds,
adapters, and evaluator settings are selected on development tasks.
Semantic evaluation uses timestamped contact sheets in a read-only,
network-free environment with access restricted to metric-specific evidence.
The held-out evaluator, its prompts, and the five reported metrics are not
used as post-SFT training signals. The metric-validation protocol is described
in Appendix~\ref{app:diagnostic-validation}, and evidence collection and
isolation within the read-only evaluation environment are detailed in
Appendix~\ref{app:harness}.

\subsection{Policy Comparison}

\begin{table}[htbp]
\caption{Policy comparison on 100 held-out tasks in the shared tool environment.
Component scores are mean$\pm$standard deviation across five runs over the
full test set. Overall is the equal-weight mean of the five component means.
Calls and generated tokens include all attempts and are normalized by
successful output minutes. Arrows indicate preferred directions, and the best
results are bold.}
\label{tab:main}
\centering
\tablefontsize
\setlength{\tabcolsep}{3.2pt}

\begin{tabular}{@{}lcccccc|cc@{}}
\toprule

& \multicolumn{6}{c}{\textbf{Task-specific performance $\uparrow$}}
& \multicolumn{2}{|c}{\textbf{PCE $\downarrow$}} \\

\cmidrule(lr){2-7}
\cmidrule(lr){8-9}
\textbf{Method} & \textbf{SCF} & \textbf{VNSM} & \textbf{TPR} &
\textbf{LHC} & \textbf{ERY} & \textbf{Overall} & \makecell{\textbf{Calls}\\\textbf{/ min}} &
\makecell{\textbf{Tokens}\\\textbf{/ min}}\\
\midrule

\rowcolor{basegroup}\multicolumn{9}{l}{\textbf{General-purpose baselines}}\\
Qwen3.5-9B & $69.8{\pm}1.6$ & $59.2{\pm}1.8$ & $79.5{\pm}1.7$ & $93.7{\pm}0.8$ & $70.4{\pm}2.1$ & 74.5 & \textbf{3.1} & 11,883\\
Qwen3.6-27B & $72.6{\pm}1.4$ & $59.7{\pm}1.6$ & $80.1{\pm}1.5$ & $95.1{\pm}0.7$ & $72.1{\pm}1.9$ & 75.9 & 3.9 & 6,742\\
Qwen3.6-35B-A3B & $71.9{\pm}1.9$ & $56.7{\pm}2.0$ & $83.4{\pm}1.8$ & $97.3{\pm}0.6$ & $79.4{\pm}2.3$ & 77.7 & 4.7 & 27,554\\
\rowcolor{uppergroup}\multicolumn{9}{l}{\textbf{Reference models}}\\
Qwen3.5-397B-A17B & $75.8{\pm}1.5$ & $64.1{\pm}1.7$ & $84.6{\pm}1.6$ & $96.2{\pm}0.8$ & $86.9{\pm}1.8$ & 81.5 & 6.0 & 15,223\\
Gemini 3.1 Pro & $80.2{\pm}1.2$ & $66.5{\pm}1.3$ & $88.2{\pm}1.2$ & $98.0{\pm}0.5$ & $90.3{\pm}1.5$ & 84.6 & 4.5 & 13,814\\
Claude Opus 5 & $87.1{\pm}1.0$ & $\mathbf{74.8}{\pm}1.1$ & $93.1{\pm}1.0$ & $98.6{\pm}0.4$ & $95.8{\pm}1.2$ & 89.9 & 11.6 & 78,763\\
\rowcolor{traingroup}\textbf{\agent (Ours)} & $\mathbf{88.4}{\pm}2.1$ & $74.6{\pm}2.3$ & $\mathbf{94.0}{\pm}2.0$ & $\mathbf{98.8}{\pm}0.5$ & $\mathbf{96.7}{\pm}1.8$ & \textbf{90.5} & \textbf{3.1} & \textbf{2,377}\\
\bottomrule
\end{tabular}
\end{table}

We first compare policies within the shared production environment to examine
the effect of task-specific training without changing the tool pipeline.
We report task performance and policy-generation cost for each evaluated policy variant
in Table~\ref{tab:main}.

\textbf{Task-specific performance.}
\agent improves the Overall Score from 74.5 to 90.5 over its Qwen3.5-9B
initialization, a gain of 16.0 points, with improvements in all five metrics.
The gains extend beyond successful delivery to script fidelity and visual
grounding, as SCF and VNSM increase by 18.6 and 15.4 points, respectively.
By comparison, the unadapted models show less uniform improvements over the
9B base policy. Qwen3.6-35B-A3B gains 9.0 points on ERY but loses 2.5 points on
VNSM, illustrating that delivery success does not necessarily imply better
visual--narration alignment. \agent also obtains higher means on all five
metrics than Qwen3.5-397B-A17B and Gemini 3.1 Pro.

\textbf{Comparison with Claude Opus 5.}
\agent achieves an Overall Score of 90.5, compared with 89.9 for Claude
Opus 5. Its mean scores are higher on four metrics and slightly lower on
VNSM, with no component differing by more than 1.3 points. Yet it uses
2,377 generated tokens per successful output minute, compared with 78,763
for Claude Opus 5, or approximately one thirty-third as many. Relative to
its base policy, \agent also reduces token usage from 11,883 while
maintaining 3.1 tool calls per successful output minute. These results
show that the task-adapted 9B \agent policy achieves mean task scores close
to those of Claude Opus 5 at substantially lower generation cost. We next
examine quality and cost in complete editing workflows with each system's native tools.

\subsection{End-to-End System Comparison}

\begin{table}[htbp]
\caption{End-to-end comparison using each system's native workflow on the
same 100 tasks. Component scores are mean$\pm$standard deviation across five
runs, and Overall is the equal-weight mean of the component means. Native
costs include all attempted runs and report wall-clock minutes and estimated
USD per successful output minute. The best results are bold.}
\label{tab:native-systems}
\centering
\tablefontsize
\setlength{\tabcolsep}{2.4pt}
\begin{tabular}{@{}lcccccc>{\columncolor{costgroup}}c>{\columncolor{costgroup}}c@{}}
\toprule
& \multicolumn{6}{c}{\textbf{Task-matched performance $\uparrow$}} &
\multicolumn{2}{c}{\cellcolor{costgroup}\textbf{Native cost $\downarrow$}}\\
\cmidrule(lr){2-7}\cmidrule(l){8-9}
\textbf{System} & \textbf{SCF} & \textbf{VNSM} & \textbf{TPR} &
\textbf{LHC} & \textbf{ERY} & \textbf{Overall} & \makecell{\textbf{Time (min)}\\\textbf{/ out min}} &
\makecell{\textbf{Cost (USD)}\\\textbf{/ out min}}\\
\midrule
\makecell[l]{Crayotter} & $81.6{\pm}1.3$ & $67.8{\pm}1.4$ & $89.1{\pm}1.3$ & $94.2{\pm}0.9$ & $91.6{\pm}1.4$ & 84.9 &  8.6 &  0.97\\
\makecell[l]{VideoAgent} & $78.4{\pm}1.5$ & $65.2{\pm}1.6$ & $86.7{\pm}1.5$ & $93.5{\pm}0.6$ & $88.7{\pm}1.7$ & 82.5 & 10.4 &  0.58\\
\rowcolor{traingroup}\textbf{\agent (Ours)} & $\mathbf{88.4}{\pm}2.1$ & $\mathbf{74.6}{\pm}2.3$ & $\mathbf{94.0}{\pm}2.0$ & $\mathbf{98.8}{\pm}0.5$ & $\mathbf{96.7}{\pm}1.8$ & \textbf{90.5} & \textbf{2.1} & \textbf{0.09}\\
\bottomrule
\end{tabular}
\end{table}

We next evaluate whether the gains persist when each editor uses its own
tools and orchestration. In Table~\ref{tab:native-systems}, we compare \agent
with Crayotter and VideoAgent using identical task inputs.

\agent achieves the highest mean score on all five metrics, with an Overall
Score of 90.5 compared with 84.9 for Crayotter and 82.5 for VideoAgent.
The improvements cover both semantic fidelity and execution quality. This pattern indicates consistent gains across content selection, temporal
planning, execution stability, and final delivery.
For example, relative to Crayotter, SCF and VNSM each improve by 6.8 points,
alongside higher plan realization, long-horizon consistency, and rendering
yield. The advantage observed in the shared environment therefore also holds
for these complete editing workflows.

These gains are accompanied by lower end-to-end costs. For each successful
output minute, \agent requires 2.1 wall-clock minutes, compared with 8.6 for
Crayotter and 10.4 for VideoAgent. The corresponding estimated monetary costs
are \$0.09, \$0.97, and \$0.58, respectively. Both measures include
unsuccessful attempts. Note that these costs reflect the complete native
workflows, rather than the efficiency of an isolated model or tool.

\subsection{Training Stage Analysis}

\begin{table}[htbp]
\caption{Cumulative training-stage analysis with a fixed Qwen3.5-9B backbone
and evaluation environment. Phase 1 includes SFT and RFT,
Phase 2 adds Guided-OPD, and Phase 3 adds \rcfdapo. Component scores
are mean$\pm$standard deviation across five runs on the 100 held-out tasks.
Overall is the equal-weight mean of the component means. The best results are
bold.}
\label{tab:training-progression}
\centering
\tablefontsize
\setlength{\tabcolsep}{3.0pt}
\begin{tabular}{@{}clcccccc@{}}
\toprule
\textbf{Phase} & \textbf{Variant} &
\textbf{SCF} & \textbf{VNSM} & \textbf{TPR} & \textbf{LHC} & \textbf{ERY} & \textbf{Overall}\\
\midrule
-- & Base & $69.8{\pm}1.6$ & $59.2{\pm}1.8$ & $79.5{\pm}1.7$ & $93.7{\pm}0.8$ & $70.4{\pm}2.1$ & 74.5\\
1 & SFT & $73.6{\pm}2.0$ & $60.5{\pm}2.2$ & $81.2{\pm}2.1$ & $94.6{\pm}1.0$ & $87.6{\pm}2.4$ & 79.5\\
1 & SFT$\rightarrow$\rft & $79.8{\pm}1.8$ & $65.1{\pm}2.0$ & $86.9{\pm}1.9$ & $95.8{\pm}0.9$ & $89.8{\pm}2.2$ & 83.5\\
2 & SFT$\rightarrow$\rft$\rightarrow$\gopd & $81.0{\pm}1.6$ & $67.4{\pm}1.8$ & $87.8{\pm}1.7$ & $96.6{\pm}0.8$ & $91.5{\pm}1.9$ & 84.9\\
3 & \textbf{\agent (Ours)} & $\mathbf{88.4}{\pm}2.1$ & $\mathbf{74.6}{\pm}2.3$ & $\mathbf{94.0}{\pm}2.0$ & $\mathbf{98.8}{\pm}0.5$ & $\mathbf{96.7}{\pm}1.8$ & \textbf{90.5}\\
\bottomrule
\end{tabular}
\end{table}

\begin{figure}[htbp]
    \centering
    \includegraphics[width=\linewidth]{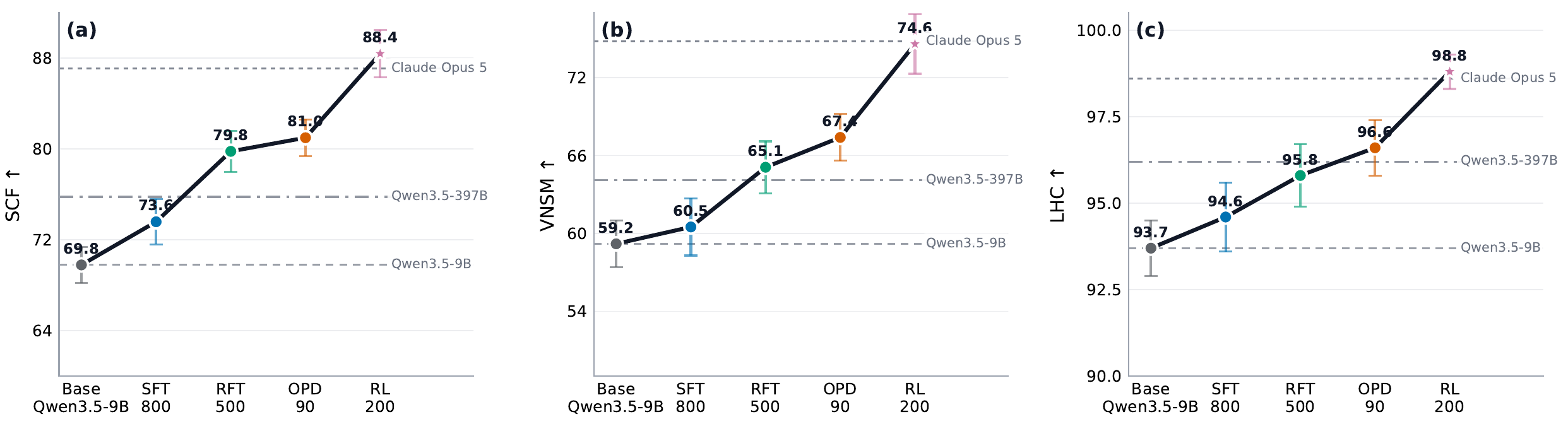}
    \caption{Cumulative training progression for (a) SCF, (b) VNSM, and (c) LHC.
    OPD and RL denote Guided-OPD and \rcfdapo, respectively. Error bars show
    standard deviations across five evaluation runs on the 100 held-out tasks.
    Horizontal lines show Qwen3.5-9B, Qwen3.5-397B-A17B, and Claude Opus 5
    under the shared tool environment. Results at these checkpoints on all
    five metrics are shown in Table~\ref{tab:training-progression}.}
    \label{fig:training-stage-progression}
\end{figure}

To examine how the training phases in Section~\ref{sec:training} contribute to the final policy,
we evaluate successive checkpoints in Table~\ref{tab:training-progression}.
The backbone, data split, tools, decoding configuration, and rendering budget
remain fixed. Each checkpoint adds the next training stage to the preceding
one, allowing us to track the cumulative gains across the three training phases.

In Phase 1, SFT raises ERY from 70.4 to 87.6, providing the largest
stage-wise improvement in delivery reliability. This is consistent with its
role in establishing valid tool use from executable demonstrations. RFT then
adds supervision from successful student trajectories and teacher-repaired
failures. Relative to SFT alone, it improves SCF by 6.2 points and TPR by 5.7
points, indicating gains in script fidelity and plan realization beyond the
initial improvement in delivery.

Building on this initialization, Phase 2 applies Guided-OPD to mixed
teacher--student rollouts. All five metrics improve at this checkpoint,
including VNSM from 65.1 to 67.4 and ERY from 89.8 to 91.5. This pattern
suggests that Guided-OPD complements trajectory-level RFT by refining decisions
at intermediate states, where the student must continue from its own partially
correct actions rather than replaying teacher-generated contexts. These results
show further gains after distillation on states reached during policy
execution, although the aggregate metrics do not isolate failure recovery.

Finally, Phase 3 optimizes production decisions with \rcfdapo. SCF and
VNSM improve by another 7.4 and 7.2 points, respectively, while Overall
increases from 84.9 to 90.5. Figure~\ref{fig:training-stage-progression}
shows the cumulative improvements in script fidelity, visual--narration
alignment, and long-horizon consistency against the reference policies.
These comparisons measure the benefit of adding training stages in this
order, rather than the independent effects of turn guidance, snapshot
matching, or individual role rewards.

\subsection{Human Evaluation}

To test whether the automatic gains carry over to human judgments, three qualified
annotators independently rate anonymized videos from the three systems in
Table~\ref{tab:native-systems} on the same 100 held-out tasks. Videos are shown individually
in random order, with scripts, source pools, and target durations fixed. The shared
tasks allow paired comparisons. Each video is rated from 1 to 5 on the six
dimensions in Table~\ref{tab:human}.

\begin{table}[htbp]
\caption{Blinded human evaluation on 100 held-out tasks with three raters. Entries
are mean 1--5 ratings (higher is better). Overall preference is a human rating,
distinct from the Overall Score.}
\label{tab:human}
\centering
\tablefontsize
\setlength{\tabcolsep}{4.5pt}
\begin{tabular}{lcccccc}
\toprule
\textbf{Method} & \makecell{\textbf{Theme}\\\textbf{fidelity}} &
\makecell{\textbf{Content}\\\textbf{richness}} &
\makecell{\textbf{Narrative}\\\textbf{coherence}} &
\makecell{\textbf{Editing}\\\textbf{smoothness}} &
\makecell{\textbf{Visual}\\\textbf{quality}} &
\makecell{\textbf{Overall}\\\textbf{preference}}\\
\midrule
Crayotter~\citep{yan2026crayottergrpb}& 3.58 & 3.42 & 3.35 & 3.48 & 3.55 & 3.48 \\
VideoAgent~\citep{zhou2026videoagent}& 3.46 & 3.60 & 3.30 & 3.42 & 3.52 & 3.46 \\
\rowcolor{traingroup}\textbf{\agent (Ours)} & \textbf{4.45} & \textbf{4.32} & \textbf{4.58} & \textbf{4.52} & \textbf{4.38} & \textbf{4.56} \\
\bottomrule
\end{tabular}
\end{table}

As shown in Table~\ref{tab:human}, WeVisCut obtains the highest mean rating
on all six dimensions, with an overall preference of 4.56 versus 3.48 for
Crayotter and 3.46 for VideoAgent. The largest
gains occur in narrative coherence (1.23 and 1.28 points), while higher editing
smoothness and visual quality indicate that the preference is not confined to
narrative structure, consistent with the automatic comparisons. Ratings are
averaged across annotators within each task. These descriptive means do not
establish statistical significance or validate individual automatic metrics
(see Appendix~\ref{app:metric-validation}).

\section{Conclusion}

We presented \agent, a system for long-form script-to-video production built
around a single 9B policy shared by the orchestrator, generator, and critic.
Our four-stage data pipeline combines editing inversion and verified replay
to recover executable supervision from finished videos. Training progresses
from executable imitation to recovery learning and reward-based decision
optimization. Our 100-task benchmark connects source scripts, editing plans,
execution traces, and final videos through five complementary metrics.
The trained policy improves all five metrics over its base model and achieves
mean scores close to those of Claude Opus 5 while using approximately one thirty-third
as many generated tokens per successful output minute. Native-system
comparisons and blinded human ratings also favor \agent. Broader template
coverage and more robust retrieval are needed to extend the system to a wider
range of editing tasks.

\FloatBarrier
\clearpage
\bibliographystyle{plainnat}
\setlength{\bibhang}{0pt}
\setlength\bibindent{0pt}
\bibliography{references}

\clearpage
\appendix

\section{Optimization and Training Details}

We use the notation of Sections~\ref{sec:agent-framework} and~\ref{sec:training} throughout: $t$ indexes role calls,
$j$ indexes tokens, and $n$ indexes training updates. The task input is $x$,
and $\pi_\theta^{(r)}$ is the shared policy conditioned on role $r$.

\subsection{Executable RFT and Guided-OPD}
\label{app:opd}

\textbf{Replay-safe RFT.}
Let $t^\star$ index the call at which the first runtime failure occurs.
A failure is recoverable when
the environment can restore the latest transaction snapshot before
$t^\star$. External side effects are executed in a sandbox or through
idempotent calls. We retain the valid prefix, provide the teacher with the
failure record and restored state, and regenerate the complete suffix.
The repaired trajectory is retained only after replay from the snapshot
passes schema, prerequisite, artifact-reference, render, duration, narration,
lint, and black-frame checks. Its quality score $Q_{\rm RFT}$ also needs to
meet the development-selected minimum $q_{\rm RFT}$:
$Q_{\rm RFT}\geq q_{\rm RFT}$. Complete student successes pass the same
admission checks. Student-only and teacher-assisted records have separate
provenance tags and fixed source--task quotas. Unrecoverable cases are kept
in a class-balanced, capped hard-case pool for later rollout sampling,
with fixed mixing probability $\eta$.

\textbf{Turn-level intervention and loss routing.}
Following Guided-OPD~\guidedopdcite, call $t$ consists of a complete role
output $z_t$ followed by its runtime observation $o_t$. At update $n$,
$q_t\sim\operatorname{Bernoulli}(\beta_n)$ selects the teacher when $q_t=1$
and the student otherwise. The selected output is executed before control
is sampled again. Both branches contribute distillation loss. We denote
the mixed rollout distribution by $\mu_n=\pi_{\theta,\phi}^{\beta_n}$,
where $\theta$ and $\phi$ are the student and frozen teacher parameters.

The models share a tokenizer. For teacher outputs, the teacher distribution
is renormalized over its $K_{\rm vocab}$ most likely tokens, giving
$\widetilde\pi_\phi^{K_{\rm vocab}}$. Student outputs use reverse KL, while
teacher outputs use this truncated forward KL:
\begin{equation}
\mathcal L_{\rm OPD}=\mathbb E_{\tau\sim\mu_n}\!\left[
\frac{1}{N_\tau}\sum_{t,j}m_{t,j}\left(
(1-q_t)D_{\rm KL}(\pi_\theta\Vert\pi_\phi)
+q_tD_{\rm KL}(\widetilde\pi_\phi^{K_{\rm vocab}}\Vert\pi_\theta)
\right)_{c_{t,j}}\right].
\label{eq:opd-detail}
\end{equation}
Here $c_{t,j}$ contains the role context $c_t$ and all tokens preceding
position $j$ in call $t$. The policies in each KL term are conditioned on
this token prefix. The mask $m_{t,j}$ is one for assistant tokens and zero
for input, padding, and runtime observations, and
$N_\tau=\sum_{t,j}m_{t,j}>0$ counts supervised tokens.
Thus teacher-only trajectories also train the student. We log retained
top-$K_{\rm vocab}$ probability mass to audit truncation. This is an
efficiency variant of Guided-OPD. Step-Level OPD instead keeps trajectories
student-controlled and queries teacher targets afterward~\stepopdcite.

Teacher intervention follows
\begin{equation}
\beta_n=\beta_{\rm start}
\frac{1+\cos(\pi\min(n/T_{\rm decay},1))}{2},
\qquad \beta_{\rm start}=0.8,
\end{equation}
where $T_{\rm decay}$ is the number of updates over which intervention
decays, set to 70\% of training. The remaining 30\% uses student-only
rollouts. Only checkpoints evaluated without teacher intervention proceed
to RL. Guided-OPD uses a 50-call horizon with early termination, while RL
uses the 16-call deployment budget. Token limits are per-call
context/response caps, and compact state updates limit history growth.

\subsection{Role- and Context-Factored DAPO}
\label{app:dapo}

\textbf{Optimization scope.}
Each trainable call records its role $r$, context $c$, output $z$, observation
$o$, and execution contract $u$. Main rollouts populate a snapshot buffer.
For an orchestrator context $c_t$, we restore the pre-decision state,
sample alternative outputs, and execute each with the same frozen
continuation policy $\pi_{\theta_{\rm old}}$, budget, and seed set.
Here $\theta_{\rm old}$ is the policy parameter snapshot frozen for the
current optimization iteration.
Nested generator and critic tokens are masked from the orchestrator loss.
Generator and critic contexts are also collected from current-policy
rollouts and resampled locally. Let $d_r$ be the context distribution for
role $r$ and $\lambda_r$ its outer loss weight. The optimized proxy is
\begin{equation}
J_{\rm proxy}=\sum_{r\in\{\rm O,G,C\}}\lambda_r
\mathbb E_{c\sim\operatorname{sg}(d_r),z\sim\pi_\theta^{(r)}(\cdot\mid c)}
[R^{(r)}(c,z)].
\label{eq:proxy-objective}
\end{equation}
Here $R^{(r)}(c,z)$ is the expected execution reward for output $z$ in
context $c$, with runtime observations averaged out.
This is a conditional call-level objective. Frozen continuations and the
sampled context distribution remain part of the proxy, so it does not
provide an unbiased gradient of the terminal production return.

\textbf{Matched comparison groups.}
For orchestrator call $t$, we fix
$\kappa_{\rm O,t}=(x,c_t,s_t,\mathcal T,B_t,\pi_{\rm cont})$:
the task input, context, restored state, tool space, remaining budget $B_t$,
and continuation policy $\pi_{\rm cont}=\pi_{\theta_{\rm old}}$.
Only the next output varies. Generator groups fix the local request,
artifact, tool, schema/version, budget, and initial state.
Critic groups also fix the defect and verification contract. Candidate
executions share paired seeds. Let $\mathcal G_\kappa$ be the group of
outputs sampled under matching conditions $\kappa$.

Features are mapped to their registered ranges: $[0,1]$, except repair
utility in $[-1,1]$. Fixed weights produce one reward per output before
group normalization. Contexts are sampled through role--task--tool quotas.
A group is accepted only if
$\operatorname{Var}_{i\in\mathcal G_\kappa}(R_i^{(r)})>\delta$,
where $\delta$ is the minimum reward-variance threshold. Otherwise it is
resampled up to a fixed cap. Low-variance contexts remain in a coverage
queue and are revisited through executable imitation. For an accepted group,
\begin{equation}
A_i^{(r)}=\operatorname{sg}\!\left[
\frac{R_i^{(r)}-\mu_\kappa^{(r)}}{\sigma_\kappa^{(r)}+\epsilon}\right],
\qquad
\mu_\kappa^{(r)}
=\frac{1}{|\mathcal G_\kappa|}\sum_{i\in\mathcal G_\kappa}R_i^{(r)}.
\label{eq:role-advantage}
\end{equation}
Here $\sigma_\kappa^{(r)}$ is the group reward standard deviation,
$\epsilon>0$ stabilizes normalization, and $\operatorname{sg}$ stops
gradients through group statistics. No normalization mixes roles or
unmatched contexts. Resampling emphasizes informative reward differences
and changes the sampled context distribution.

\textbf{Role-specific rewards.}
For candidate $i$, let $z_i$ be its role output and $o_i$ its execution
feedback. An orchestrator output is scored by continuing from snapshot $t$.
We measure progress through validated milestones:
\begin{align}
\Phi(s)&=\frac{\sum_k\omega_k I_k(s)}{\sum_k\omega_k},\qquad
r_\ell^{\rm prog}=\gamma\Phi(s_{\ell+1})-\Phi(s_\ell),\notag\\
F_{i,t}^{\rightarrow}&=(1-\epsilon_p)
\operatorname{clip}\!\left(
\sum_{\ell\geq t}\gamma^{\ell-t}r_\ell^{\rm prog},0,1\right).
\label{eq:progress-potential}
\end{align}
Here $I_k(s)\in\{0,1\}$ indicates whether unique milestone $k$ holds in
state $s$, $\omega_k>0$ is its fixed weight, $0<\gamma\leq1$ is the discount
factor, and $\ell$ indexes subsequent calls until branch $i$ terminates. Repetition adds no
milestone credit, and later invalidation removes it. The positive margin
$0<\epsilon_p<1$ keeps failed-branch progress below the delivery tier.
The operator $\operatorname{clip}(\cdot,0,1)$ bounds its argument to $[0,1]$.

The rewards are
\begin{align}
R_i^{\mathrm O}&=g_{{\rm del},i}
[\alpha+(1-\alpha)Q_{{\rm end},i}]
+(1-g_{{\rm del},i})\alpha F_{i,t}^{\rightarrow}
-\lambda_p p_i,\notag\\
R_i^{\mathrm G}&=\sum_k w_k^{\mathrm G}f_k^{\mathrm G}(c,z_i,o_i),\notag\\
R_i^{\mathrm C}&=w_d f_{{\rm diag},i}
+w_r U_{\rm repair}(d_i)+w_v f_{{\rm verify},i}
-w_u f_{{\rm unnecessary},i}.
\label{eq:role-reward}
\end{align}
The delivery indicator $g_{{\rm del},i}$ is one only if termination, schema,
prerequisite, artifact, and loop checks pass and final quality
$Q_{{\rm end},i}\geq q_{\min}$, where $q_{\min}$ is the minimum delivery
quality. Otherwise it is zero. A legal stop below this threshold is
premature, recorded by the indicator $p_i\in\{0,1\}$.
The parameter $\alpha\in[0,1]$ weights delivery and progress credit
relative to final quality, and $\lambda_p\geq0$ weights the premature-stop penalty.

Final quality combines a rule score and a sampled render audit:
$Q_{\rm end}=(1-w_a)Q_{\rm rule}+w_aQ_{\rm audit}$,
where $Q_{\rm rule},Q_{\rm audit}\in[0,1]$ and $w_a\in[0,1]$ is the fixed
audit weight. The six rule features are plan
coverage, spotlight-decoration coverage, inspection coverage, plan hygiene,
completion, and efficiency, weighted 0.353, 0.186, 0.179, 0.144, 0.081, and
0.057. The render audit scores visual impact and clarity. Thus these six
weights apply to rule features, with the audit blended separately.

Generator features $f_k^{\mathrm G}$ measure schema validity, grounding,
requirement coverage, and downstream usability using deterministic checks
and a frozen tool-specific rubric. All features are set to zero for an
invalid schema. Critic features $f_{{\rm diag},i}$ and
$f_{{\rm verify},i}$ score diagnosis and verification,
and $f_{{\rm unnecessary},i}$ penalizes unnecessary repair.
The coefficients $w_k^{\mathrm G},w_d,w_r,w_v,w_u$ are fixed feature weights.
Here $d_i$ is the diagnosis provided by critic output $z_i$. For a diagnosis
$d$, controlled repair utility is
\begin{equation}
U_{\rm repair}(d)=\frac{1}{|\Xi|}\sum_{\xi\in\Xi}
\left[Q(A_{\rm fix}(d;\xi))-Q(A_{\rm fix}(d_0;\xi))\right],
\label{eq:critic-utility}
\end{equation}
where $A_{\rm fix}$ is a frozen repair executor with identical tool version
and budget, $d_0$ is an empty diagnosis, $\Xi$ is a paired seed set, and
$Q$ maps repair results to a fixed quality score in $[0,1]$. Thus utility measures the
improvement over the empty-diagnosis baseline under controlled execution.
The training judge, rubric, prompt, and version hash are frozen and
independent of the held-out evaluator. Rule and judge scores are logged
separately.

\textbf{DAPO update.}
Let $\mathcal B_r$ be the candidate batch for role $r$, and
$\mathcal M_i^{(r)}$ the model-authored token positions in output $z_i$.
For token $j$, the policy probability ratio is
$\rho_{i,j}=\pi_\theta(z_{i,j}\mid c_i,z_{i,<j})/
\pi_{\theta_{\rm old}}(z_{i,j}\mid c_i,z_{i,<j})$,
where $c_i$ is the candidate's input context.
We use DAPO's asymmetric Clip-Higher interval~\citep{yu2025dapo}, with
$\varepsilon_{\rm low}=0.20$ and $\varepsilon_{\rm high}=0.28$:
\begin{align}
\ell_{i,j}^{(r)}&=\min\!\left(
\rho_{i,j}A_i^{(r)},
\operatorname{clip}(\rho_{i,j},0.8,1.28)A_i^{(r)}\right),\notag\\
\mathcal J_{\rm RCF\text{-}DAPO}
&=\sum_r\lambda_r\,
\mathbb E\!\left[
\frac{\sum_{i\in\mathcal B_r}\sum_{j\in\mathcal M_i^{(r)}}
\ell_{i,j}^{(r)}}
{\sum_{i\in\mathcal B_r}|\mathcal M_i^{(r)}|}
\right].
\label{eq:rcf-dapo-objective}
\end{align}
Token losses are aggregated within each role before applying the outer
weights $\lambda_{\rm O}=\lambda_{\rm G}=\lambda_{\rm C}=1/3$.
The old policy is synchronized once per optimization iteration.
We use neither dual clipping nor a loss-side KL term.

For generated token count $L$, let $L_{\rm soft}<L_{\rm cap}$ denote the
no-penalty threshold and generation hard cap. For non-truncated responses,
soft overlong shaping is
\begin{equation}
p_{\rm len}(L)=
\begin{cases}
0,&L\leq L_{\rm soft},\\
-\lambda_{\rm len}\dfrac{L-L_{\rm soft}}{L_{\rm cap}-L_{\rm soft}},
&L_{\rm soft}<L<L_{\rm cap}.
\end{cases}
\label{eq:overlong-shaping}
\end{equation}
We set $L_{\rm soft}=L_{\rm cap}-2{,}048$ and the length-penalty weight
$\lambda_{\rm len}=0.1$.
Responses reaching the hard cap are replaced before an eight-candidate
group is formed. If the resampling cap is reached first, the group is
dropped. Truncated responses enter neither rewards, group statistics,
nor policy loss. Role-wise gradient norms and pairwise cosine similarities
are logged, since outer loss weights control scale but do not eliminate
shared-parameter gradient conflict.

The registered post-training configuration is shown in
Table~\ref{tab:training-config}.

\begin{table}[H]
\caption{Registered post-training configuration.}
\label{tab:training-config}
\centering
\small
\begin{tabularx}{\linewidth}{@{}lX@{}}
\toprule
Stage & Key configuration \\
\midrule
Guided-OPD & Qwen3.5-9B student with a Qwen3.5-397B-A17B-FP8 teacher and
$K_{\rm vocab}=50$. Per-call prompt/response caps: 12,288/4,096 tokens.
Batch 16, 90 steps, $\beta_{\rm start}=0.8$ decaying over 70 steps, and
learning rate $10^{-6}$. At most 50 assistant turns with early termination. \\
RCF-DAPO & Per-call prompt/response caps: 24,576/12,288 tokens.
Train/generation/mini-batch: 48/72/24, matched call group size 8, and separate
O/G/C batches. Stratified dynamic sampling with coverage replay.
Clip $[0.8,1.28]$, temperature 0.9, 16-turn deployment horizon, learning rate
$10^{-6}$, LoRA rank/alpha 32/64, and no KL penalty. \\
\bottomrule
\end{tabularx}
\end{table}

\section{Metrics and Evidence Protocol}
\label{app:metrics}

The five component metrics lie in $[0,100]$ and are reported individually,
alongside the descriptive Overall Score defined below. Component claims, shot
IDs, matching rules, thresholds, and denominators are frozen before test
outputs are inspected.

\textbf{Script Content Fidelity (SCF).}
Let $c_1,\ldots,c_m$ be the claims in source script $S$, and
$s_1,\ldots,s_k$ the claims in output narration $\mathcal V$.
The entailment indicator $E(a,B)\in\{0,1\}$ tests whether text $B$ supports claim $a$,
and $\tau_K$ is the Kendall order correlation of matched source and output claims.
We compute
\begin{align}
\mathrm{Cov}&=m^{-1}\sum_{i=1}^{m}E(c_i,\mathcal V),&
\mathrm{Faith}&=k^{-1}\sum_{j=1}^{k}E(s_j,S),\notag\\
\mathrm{Ord}&=(1+\tau_K)/2,&
\mathrm{SCF}&=100(0.40\mathrm{Cov}+0.35\mathrm{Faith}+0.25\mathrm{Ord}).
\label{eq:scf}
\end{align}
SCF is valid for at least five source claims and one narration unit.

\textbf{Visual--Narration Semantic Match (VNSM).}
Let $\mathcal S$ contain the registered shots with a valid evaluator response.
For each shot $s$, four uniformly sampled frames form montage $F_s$, paired
with concurrent narration $T_s$. The held-out Claude Opus 5 evaluator assigns
an anchored grade $g_s=g(F_s,T_s)\in\{1,2,3,4\}$ (Poor, Fair, Good,
Excellent) using the rubric below. We rescale grades to $[0,100]$ and aggregate
\begin{equation}
\mathrm{VNSM}=|\mathcal S|^{-1}\sum_{s\in\mathcal S}\frac{100(g_s-1)}{3},
\qquad |\mathcal S|\geq5.
\label{eq:vnsm}
\end{equation}

\textbf{Timeline Plan Realization (TPR).}
Planned and realized shots are joined by frozen shot ID:
\begin{align}
R_{\rm pres}&=|P\cap A|/|P|,\notag\\
R_{\rm dur}&=|P\cap A|^{-1}\!\sum_{s\in P\cap A}
\exp\!\left[-\frac{|D_s^{\rm plan}-D_s^{\rm act}|}
{\max(0.5,0.3D_s^{\rm plan})}\right],\notag\\
R_{\rm ord}&=\mathrm{LCS}(P,A)/\max(|P|,|A|),\notag\\
\mathrm{TPR}&=100(0.30R_{\rm pres}+0.40R_{\rm dur}+0.30R_{\rm ord}).
\label{eq:tpr}
\end{align}
We set $R_{\rm dur}=0$ when $P\cap A=\varnothing$.

\textbf{Long-Horizon Consistency (LHC).}
Shot health averages render success, audio--clip duration agreement, and the
rescaled VNSM grade:
\begin{equation}
q_s=\frac{1}{3}\!\left[\mathbf1(s\text{ rendered})+
\exp\!\left(-\frac{|D_s^{\rm audio}-D_s^{\rm clip}|}
{\max(0.5,0.3D_s^{\rm audio})}\right)+\frac{g_s-1}{3}\right].
\end{equation}
Unavailable $g_s$ is omitted and the observable terms are renormalized. For
the first and last thirds $H,T$ and canonical asset digests $\mathcal A$,
\begin{align}
R_{\rm ret}&=\operatorname{clip}(\bar q_T/\bar q_H,0,1),&
R_{\rm nov}&=|\operatorname{unique}(\mathcal A)|/|\mathcal A|,\notag\\
\mathrm{LHC}&=100(0.60R_{\rm ret}+0.40R_{\rm nov}).
\label{eq:lhc}
\end{align}
LHC requires $N\geq12$ and $\bar q_H\geq0.5$.

\textbf{End-to-End Rendering Yield (ERY).}
\begin{equation}
\mathrm{ERY}=\frac{100}{N_{\rm total}}\sum_c\mathbf1[\mathrm{ok}(c)],\quad
\mathrm{ok}=\mathrm{exists}\wedge\mathrm{decodable}\wedge
[D^{\rm act}\geq0.5D^{\rm plan}]\wedge\mathrm{hasNarration}.
\label{eq:ery}
\end{equation}
\textsc{hasNarration} requires non-silent speech covering at least half of the
planned narration duration. Background music does not qualify. Timeouts,
crashes, and unstarted tasks remain in the denominator.

\textbf{Overall Score (descriptive summary).}
For each model or system within a comparison track, let $\overline M_k$
denote its reported mean for component $k$. We add a post-hoc descriptive
summary of these means:
\begin{equation}
\mathrm{Overall}=\frac{1}{5}\sum_{k\in\{\mathrm{SCF,VNSM,TPR,LHC,ERY}\}}
\overline M_k.
\end{equation}
Each component receives weight $0.2$. These weights are not fitted to model
rankings or human preferences. The shared $[0,100]$ scale makes this a transparent
macro-average, but does not make the components independent or equally
calibrated: LHC already incorporates VNSM in shot health. Overall therefore
summarizes task performance rather than serving as a separately validated
measure of perceptual video quality. It is distinct from the human-rated
overall preference in Table~\ref{tab:human}.

The displayed scores are computed from the reported, rounded component means
and rounded to one decimal place. Each component retains its own eligibility
rules and denominator. Overall does not imply a common per-task population.
If any component mean is unavailable, Overall is N/A rather than an average
over the remaining metrics. Costs are excluded. We report no Overall standard
deviation, confidence interval, or significance test: these require aligned
run-level results or the cross-metric covariance, which cannot be recovered
from the component standard deviations alone. All five components remain
visible so that the summary does not conceal trade-offs.

\textbf{Production Cost Efficiency (PCE).}
For tool calls or generated tokens $X$,
\begin{equation}
\mathrm{PCE}_X=\frac{\sum_{c\in\mathcal C_{\rm all}}X(c)}
{\sum_{c\in\mathcal C_{\rm ok}}D_c^{\rm act}/60}.
\label{eq:pce}
\end{equation}
PCE is lower-is-better and is never combined with quality. Eligibility counts
and validity conditions are recorded alongside metric reporting. Conditional
means use valid tasks, whereas unconditional means assign zero to system
failures. Harness or provider outages remain missing rather than being
converted into low-quality scores.

\subsection{Diagnostic Validation}
\label{app:diagnostic-validation}
\label{app:metric-validation}

We apply three strengths of metric-targeted corruption: claim edits (SCF),
time-local visual substitution (VNSM), shot removal/duration/order changes
(TPR), tail degradation or asset repetition (LHC), and delivery failure (ERY).
For each metric--corruption pair we report the paired drop, bootstrap 95\%
interval, severity monotonicity, and the $5\times5$ sensitivity matrix. Human
pairwise validation is metric-specific and blinded. VNSM additionally reports
cross-family judge agreement and human correlation. Agreement below
Krippendorff's $\alpha=0.6$ prevents a standalone VNSM claim.

\subsection{Evidence-Seeking Harness}
\label{app:harness}

At most three $4\times4$ contact sheets are generated per video with sampling
step
\begin{equation}
\Delta=\max\!\left(1,\Delta_{\rm user},
\left\lceil N/(16S_{\max})\right\rceil\right),\qquad S_{\max}=3.
\end{equation}
Tiles carry source-frame and timestamp labels, while a sidecar stores FPS,
duration, sampling step, and sheet count. Representative inputs are shown in
Appendix~\ref{app:visual-inputs}. Read-only mounts isolate the evidence available
to each metric. SCF sees script and narration, while VNSM sees aligned narration
and rendered frames/clips. TPR sees plan and realized timeline. LHC sees ordered
health inputs and asset digests. ERY/PCE see delivery probes and accounting.
The network-free harness may call only bounded FFmpeg (editing and probing),
PySceneDetect (scene detection), Whisper (ASR), PaddleOCR (OCR), frozen
detectors, and manifest/diff readers. Tools expose evidence but never scores,
and every call is archived.

\subsection{Held-Out VNSM Evaluator Prompt}

\begin{figure}[H]
    \centering
    \includegraphics[width=\linewidth]{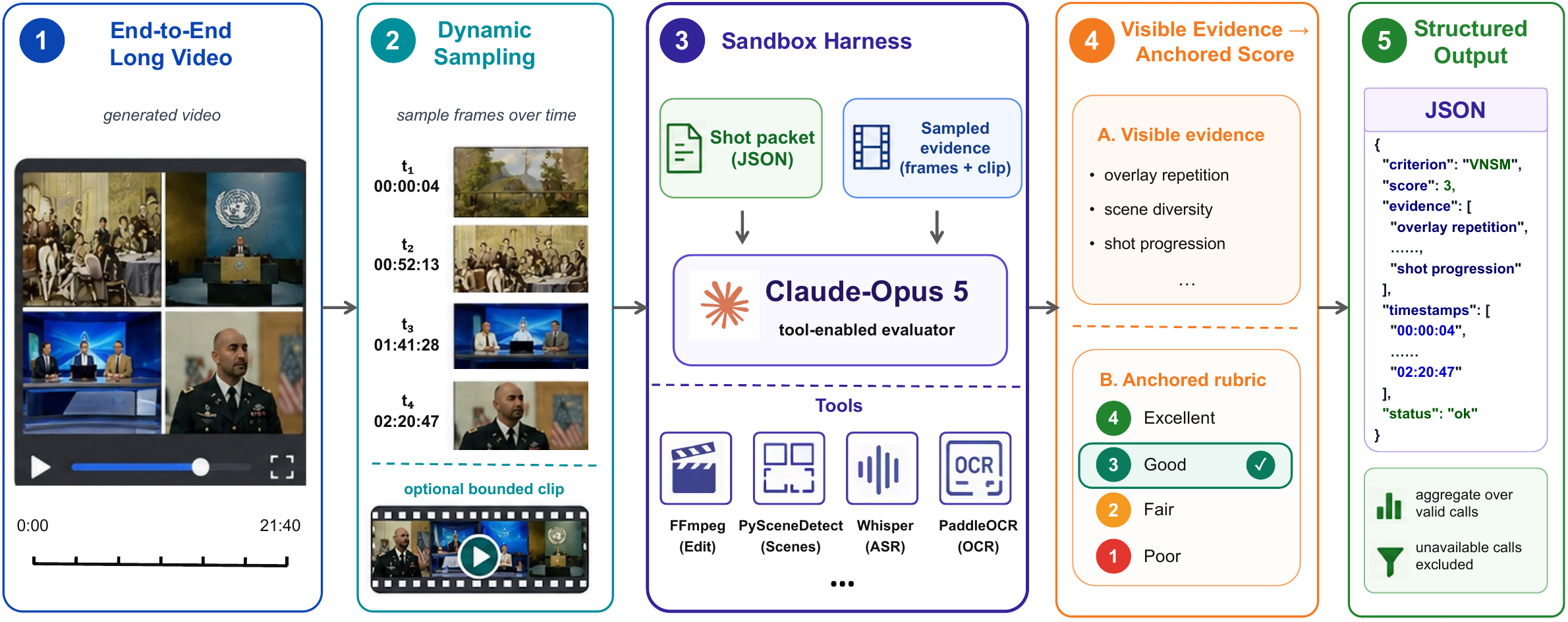}
    \caption{Auditable VNSM evaluator path. A shot packet binds the narration
    window, timestamps, video and manifest digests, sampler/prompt versions,
    judge snapshot, and decoding configuration. The frozen sampler exposes
    four timestamped frames and, only when motion is necessary, one bounded
    clip. The held-out Claude Opus 5 evaluator must first record visible
    evidence, then choose one anchored grade and emit schema-validated JSON.
    Aggregation excludes
    unavailable calls rather than converting infrastructure failure into a
    zero-quality judgment.}
    \label{fig:vnsm-harness}
\end{figure}

The evaluator is called independently for each registered shot. The
complete version tuple---prompt, rubric, model snapshot, decoding parameters,
sampler, and JSON schema---is frozen after development-only calibration and
before any test output is rendered. Figure~\ref{fig:vnsm-harness} visualizes
the exact evidence path. The four image tiles are frames sampled by the
evaluation harness rather than illustrative replacements.

\begin{promptbox}[Held-Out VNSM Evaluator: System Contract]{wedirectorblue}
\textbf{Role.} You are an evidence-seeking evaluator for one registered metric.
Evaluate only visual--narration semantic correspondence; do not grade overall
video quality, aesthetics, editing smoothness, source fidelity, tool use, or
hidden production intent.

\textbf{Immutable inputs.} A shot packet (JSON) with \texttt{criterion=VNSM},
case and shot IDs, the narration text active over the registered shot
interval, video and manifest SHA-256 digests, sampler and prompt versions, and
model snapshot and decoding configuration; and sampled evidence comprising four
chronologically ordered frames $t_1$--$t_4$ with source timestamps
(\texttt{hh:mm:ss}) and, on request, one bounded clip. No plan, retrieval
query, similarity score, reference video, method name, or training reward is
supplied.

\textbf{Procedure.}
\begin{enumerate}[leftmargin=17pt,nosep]
\item Verify that all four frames, their timestamps, and the narration window
are present and internally consistent. Otherwise set \texttt{status} to one
registered failure code.
\item Decompose the narration into its central visually testable units:
referent, setting, attribute, action, and event. Do not require support for
abstract discourse that cannot be depicted.
\item Inspect every frame in chronological order and record two to four
time-local visible observations before assigning a score. Each observation
must be paired with one supplied frame timestamp.
\item For each central unit, decide whether it is directly supported, merely
generic/topical, absent, or contradicted. One matching object cannot override
a contradictory action or event.
\item Request the single bounded clip only when motion or temporal order is
necessary to distinguish an action. The clip may cover only the registered
shot interval and cannot introduce outside context.
\item Select exactly one anchored score whose full anchor is satisfied. When
evidence falls between anchors, choose the lower anchor; never interpolate or
invent a new score.
\end{enumerate}

\textbf{Evidence rules.} A generic presenter, publisher card, or decorative
overlay alone cannot score above 2 (Fair). OCR text counts as visual support
only when the narration is explicitly about displayed text. A correct setting
with the wrong central action is at most 2 (Fair); a direct contradiction
scores 1 (Poor). Evidence outside the registered narration interval is
inadmissible. Never invent a timestamp or infer content absent from the
supplied frames or bounded clip. Missing input, decoding failure, or
harness/provider failure sets \texttt{status} to a failure code; it is missing
evidence, not a Poor score.

\textbf{Output.} Return strict JSON only, with no markdown or free-form text.
\end{promptbox}

\begin{promptbox}[Held-Out VNSM Evaluator: Anchors and Output]{wedirectorgreen}
\textbf{Anchored rubric.}
\begin{itemize}[leftmargin=17pt,nosep]
\item \textbf{4 --- Excellent (direct and complete)}: every central visually
testable referent/action/event is directly visible in the matching interval,
with no material contradiction.
\item \textbf{3 --- Good (specific but partial)}: the correct subject and scene
are visible and the central referent or action is substantially supported, but
one important attribute, phase, or consequence is not verifiable.
\item \textbf{2 --- Fair (topical or generic)}: the imagery is in the correct
broad topic or setting, but the central action, relation, or specific referent
is not visually verified. Generic B-roll belongs here even when aesthetically
apt.
\item \textbf{1 --- Poor (atmospheric, absent, or contradictory)}: only color,
mood, or a weak association relates to the narration, or the imagery is
irrelevant, shows the wrong subject/action/event, or materially contradicts
the narration.
\end{itemize}

\textbf{Required JSON fields.}
\begin{verbatim}
{
  "criterion": "VNSM",
  "score": 1|2|3|4,
  "evidence": ["<visible observation>", ...],
  "timestamps": ["<hh:mm:ss>", ...],
  "status": "ok"
}
\end{verbatim}
For \texttt{status="ok"}, \texttt{score} is an allowed anchor, and
\texttt{evidence} contains 2--4 distinct visible observations, each aligned
with one entry of \texttt{timestamps} drawn from the supplied frames.
Otherwise, \texttt{score} is null, \texttt{evidence} and \texttt{timestamps}
are empty, and \texttt{status} is exactly one of \texttt{missing\_input},
\texttt{digest\_mismatch}, \texttt{sampler\_failure},
\texttt{provider\_failure}, \texttt{decode\_failure}, or
\texttt{schema\_failure}. Case and shot IDs, digests, and version IDs are
bound by the shot packet rather than repeated in the output.
\end{promptbox}

\textbf{Deterministic validation and retry.}
The evaluator runs at temperature 0 with the frozen decoding parameters shown
in the packet. A deterministic validator checks the \texttt{criterion} and
\texttt{score} values, the \texttt{status}-dependent fields, timestamp
membership, evidence count, and the match between each response and its shot
packet. An invalid response receives one schema-only repair request containing
the validation errors but no new visual evidence. A second invalid response is
recorded as unavailable with \texttt{schema\_failure}. No human edits the score
or evidence.

\textbf{Aggregation and claim gate.}
The held-out Claude Opus 5 evaluator judges each registered shot once.
Video-level VNSM is the unweighted mean of rescaled grades over shots with
\texttt{status="ok"} and is reported only when $|\mathcal S|\geq5$.
Unavailable calls are excluded rather than scored. A second held-out VLM family
independently re-judges the same shots solely to audit agreement, and its grades
do not enter VNSM. We report coverage, eligible-shot count, the two
family-specific means, rank correlation, and Krippendorff's $\alpha$. Agreement
below $\alpha=0.6$ blocks a standalone VNSM claim. Before the prompt is frozen,
a disjoint 50-shot development set is used for blinded human correlation and
controlled corruption tests (wrong subject, wrong action, temporal shift, and
generic B-roll). Test tasks never participate in calibration.

\textbf{Training/evaluation separation.}
The training-side VLM-Judge is a sampled spotlight audit, used only for reward
features and limited to \texttt{visual\_impact} and \texttt{clarity}: the
former checks whether the animation is visually engaging in layout and
aesthetics, while the latter checks whether motion and text convey information
without distraction or layout errors.  The held-out VNSM evaluator and audit
family instead use final rendered evidence, separately versioned models,
isolated credentials, and prompts unavailable to the training jobs.

\section{Experimental Protocol and Reporting}
\label{app:experiments}

\subsection{External Baseline Contract}

Task-matched native systems receive the same frozen script, source pool,
duration, aspect ratio, and budget while retaining their released orchestration
and tools. Adapters may translate serialization and export native artifacts,
but may not retrieve assets, rewrite narration, invent plans, add manual edits,
or perform unavailable repairs. We record the plan, call log, artifacts, final
video, failure code, time, and cost. A missing native artifact makes its
trace-dependent metric N/A. Systems requiring semantic adapter work enter the
task-adjacent track and are excluded from the headline matched ranking.
The five component-metric entries average five independent runs over the same
100 held-out tasks, with $\pm$ denoting the standard deviation across runs. Overall
Score instead averages the five reported component means and is shown without
an uncertainty estimate. It is N/A when any component mean is unavailable.
Missing components are neither imputed nor omitted from the denominator.

\begin{table}[H]
\caption{Baseline task-contract audit. Systems satisfying every required
contract element enter the task-matched native-system comparison.}
\label{tab:baseline-contract}
\centering
\tablefontsize
\setlength{\tabcolsep}{3.2pt}
\begin{tabular}{@{}llccccc@{}}
\toprule
\textbf{System} & \textbf{Paradigm} & \makecell{\textbf{Fixed}\\\textbf{narration}} &
\makecell{\textbf{Shot}\\\textbf{plan}} &
\makecell{\textbf{Trace /}\\\textbf{repair}} &
\makecell{\textbf{Fixed}\\\textbf{sources}} &
\makecell{\textbf{Matched}\\\textbf{native track}} \\
\midrule
Crayotter     & Artifact-centric multi-agent   & \cmark & \cmark & \cmark & \cmark & \cmark \\
VideoAgent    & Large agent library            & \cmark & \cmark & \cmark & \cmark & \cmark \\
CutClaw       & Specialized multi-agent        & \xmark & \cmark & \cmark & \cmark & \xmark \\
CineAgents    & Design-and-compose agents      & \xmark & \cmark & \cmark & \cmark & \xmark \\
DirectorAgent & Script-to-video generation     & \xmark & \cmark & \xmark & \xmark & \xmark \\
\bottomrule
\end{tabular}
\end{table}

\subsection{Plan--Render Visualization Protocol}
\label{app:visualization}

Qualitative examples are selected from registered length, domain, and failure
strata before method identities are revealed. The native-system panel fixes
rows to Crayotter and \agent and uses one common narration and source pool.
Each row shows five rendered keyframes at identical absolute timestamps and a
compact evidence strip restricted to visible properties. Missing frames remain
visibly missing and an invalid delivery retains its terminal failure stage.
Additional systems, when available under the same contract, appear in separate
panels rather than changing the frozen comparison. Visualizations explain the
registered metrics but do not re-score or reorder examples.

\subsection{Depth and Real-Tool Diagnostics}

For a subset of tasks, the evaluation protocol injects a registered late-stage
fault and records the repair branch $\tau^+$. The shared case ID links the
script, shot plan, original execution, delivered video, and changes made during
repair, allowing recovery behavior to be inspected alongside normal production.

For horizon $H_i$ and first failure turn $t_i^{\rm fail}$,
\begin{equation}
\operatorname{FailRate}@k=
\frac{\sum_i\mathbf1[H_i\geq k]\mathbf1[t_i^{\rm fail}=k]}
{\sum_i\mathbf1[H_i\geq k]}.
\end{equation}
Completion and repair at depth $k$ use the same eligible denominator. The
proxy--real gap is a paired mean between frozen-fixture replay and production
execution. Unmatched or failed runs are retained in reporting. These depth and
proxy--real diagnostics complete the registered protocol. The experiment tables
report the five registered metrics and the descriptive Overall Score.

\section{A Representative Autonomous Tool-Use Flow}
\label{app:tool-flow}

This appendix illustrates the inference-time behavior of \agent with one
complete production trace. The purpose is not to expose the full generated
screenplay or every JSON payload, but to show the closed-loop decision pattern
used by the policy: it first builds a long-horizon production plan, then
executes real tools, observes typed runtime returns, repairs bounded failures,
satisfies missing prerequisites, and terminates only after the video has
actually been delivered.

The raw trace is stored as \path{agent_trace.jsonl} and the delivery record as
\path{stitch_manifest.json}. The transcript below is a faithful, translated
rendering of that trace, organized as
\begin{gather*}
\textsc{Thinking}
\;\rightarrow\;
\textsc{Observation}
\;\rightarrow\;
\textsc{Tool Call}\\
\rightarrow\;
\textsc{Runtime Return}
\;\rightarrow\;
\textsc{Next Decision}.
\end{gather*}
The \textsc{Thinking} field reports the deliberation the policy emits before
committing to a tool action. It is part of the recorded action interface rather
than hidden reasoning: every claim it makes about the world is either grounded
in a previous typed return or is falsified by the next one, which is precisely
what makes the trace usable as training supervision.

\begin{promptbox}
  [{Representative Autonomous Trace: Plan--Act--Repair--Deliver}]
  {wedirectorblue}
\sloppy

\textbf{Run card.}
The run identifier is
\path{aicut_20260824_152401_9f7f9d}.
The input is a fixed long-form narration about aircraft science and
engineering, tokenized by the runtime into 51 semantic units
(\texttt{u01}--\texttt{u51}). The finished project contains 6 screenplay
segments, 30 realized shots, 16 spotlight shots, and 3 background-music tracks.
The policy runs for 16 tool turns over 31 minutes of wall-clock time and
delivers a 395.0-second video:

\begin{center}
\path{final_video_with_bgm_20260824_155554.mp4}
\end{center}

\tracestep{1}{Build the high-level production plan}

\tracefield{Thinking}
I hold nothing but the narration. Every downstream tool in this runtime is
gated on structure that does not exist yet, so the only admissible first action
is the one that turns free text into a typed screenplay. Two choices matter
here. First, granularity: the narration runs several hundred seconds, and a
segment count near six keeps each act long enough to carry an argument while
still giving the shot planner meaningful boundaries to respect. Second, I
should request the global music style now rather than at delivery time, because
the stitcher will need a music query long after the narration's tone is still
in context, and recovering that tone later from rendered shots would be lossy.
I also pin narration order and full-script coverage, since any dropped span
becomes a semantic hole that no later repair tool is able to detect.

\tracefield{Observation}
Only the raw narration is available. No shot plan, assets, audio, templates,
rendered paths, or delivery artifacts exist.

\tracefield{Call}
\begin{lstlisting}[style=toolcall]
Tool_SegmentArticle(
    num_segments_target=6,
    segment_granularity="medium",
    preserve_narration_order=True,
    cover_full_script=True,
    need_music_style=True
)
\end{lstlisting}

\tracefield{Return}
\texttt{status=success}. The runtime returns a six-segment outline --- a
prologue (``Looking Up in Wonder''), four core acts (``A Precision Airborne
Ecosystem'', ``Extreme Lightweighting and Flexible Wisdom'', ``Eyes of Technology
and Invisible Guardians'', ``From Adventure to Everyday''), and a coda (``A Love
Letter to the Sky'') --- together with a per-segment narration script, a
planner strategy, and the global music style: ``epic orchestral, piano
lead, strings underneath, tempo moving from slow to fast, grand and warm''.

\tracefield{Next decision}
Convert the outline into executable shot-level production cards.

\tracestep{2}{Expand the plan into shot-level production cards}

\tracefield{Thinking}
The segment outline is an argument, not a shot list. Nothing downstream can
consume it: retrieval needs a visual query per shot, synthesis needs a vocal
script per shot, and the renderer needs durations and identifiers. I set the
shot target well above what I expect to keep, because the planner is allowed to
merge and drop but not to invent coverage, and an over-request costs nothing at
this stage while an under-request silently truncates the tail of the script. I
forbid repeated visual queries, since duplicate queries collapse to the same
retrieved asset and produce visible repetition that no per-shot quality check
will flag --- the failure only becomes obvious in the stitched cut, which is
far too late to fix cheaply. I also want spotlight marks assigned here rather
than later, because the decoration and rendering budget is the scarcest
resource in the run and I would rather allocate it from a global view of the
script than greedily, shot by shot.

\tracefield{Observation}
A segment-level plan exists, but the runtime still lacks shot IDs, visual
queries, durations, spotlight marks, and coverage records.

\tracefield{Call}
\begin{lstlisting}[style=toolcall]
Tool_PlanShots(
    mode="plan_shots_all",
    num_shots_target=50,
    planning_policy={
        "preserve_narration_order": True,
        "avoid_repeated_visual_query": True
    }
)
\end{lstlisting}

\tracefield{Return}
\texttt{status=success}. The planner realizes 30 shots spanning 397.4 planned
seconds, distributed 2/2/2/7/10/7 across the six segments, and marks 16 of them
as spotlight shots. The union of \texttt{covered\_unit\_ids} is
\texttt{u01}--\texttt{u51}, i.e. semantic coverage is complete with no
uncovered narration unit.

\tracefield{Next decision}
Materialize the planned shots with real assets and narration audio.

\tracestep{3}{Execute batched visual retrieval and narration synthesis}

\tracefield{Thinking}
Retrieval and text-to-speech are the two slowest operations in the pipeline and
they are independent per shot, so issuing them one shot at a time would waste
most of the run's wall-clock budget on serialization. I dispatch all 30 shots
in a single batch. This is also the first turn where the plan meets the real
asset library, which means it is the first turn where my assumptions can be
falsified: a visual query that reads well as prose may still retrieve nothing
usable. I therefore care less about the average match score in the return than
about whether any shot comes back with an empty candidate set, because a
partial batch would force me to choose between re-querying and accepting a hole
in the cut.

\tracefield{Observation}
The shot plan is complete, but no media artifacts exist.

\tracefield{Call}
\begin{lstlisting}[style=toolcall]
Tool_BatchVisualAudio(
    shot_ids=["1", "2", ..., "30"]
)
\end{lstlisting}

\tracefield{Return}
\texttt{status=success} for all 30 shots. Every shot receives a clip path, a
source duration, a match score, and a non-empty candidate set --- for example
shot~2 retrieves at \texttt{match\_score}~$=0.592$ from 12 candidates and
shot~5 at $0.611$ from 12 candidates, while the weakest, shot~1, still returns
$0.452$ from 5 candidates. All shots report
\texttt{playlist\_ready=true}. The batch takes roughly 14 minutes, the single
most expensive turn in the run.

\tracefield{Next decision}
Spend the decoration and rendering-template budget only on spotlight shots.

\tracestep{4}{Decorate the spotlight shots}

\tracefield{Thinking}
Decoration is what separates a slideshow from an edited video, but it is also
the most expensive thing I can ask for, both in template rendering and in the
downstream quality checks it triggers. The 16 spotlight marks from step~2 are
exactly the allocation decision I already made, so I honor them and set
\texttt{spotlight\_only} rather than re-deciding now with less context than the
planner had. I ask for slot verification in the same call, because a template
whose slots are filled but semantically wrong is worse than no template at
all: it renders cleanly, passes structural checks, and puts an incorrect
caption on screen.

\tracefield{Observation}
Shot assets and narration audio exist for all 30 shots, and the 16 spotlight
IDs are known from the plan.

\tracefield{Call}
\begin{lstlisting}[style=toolcall]
Tool_BatchDecorateSpotlights(
    shot_ids=["1","2","5","6","7","10","11","13",
              "14","16","19","21","22","23","26","28"],
    template_policy={
        "select_template": True,
        "fill_slots": True,
        "verify_slots": True,
        "spotlight_only": True
    }
)
\end{lstlisting}

\tracefield{Return}
\texttt{status=success}, \texttt{decorated\_count=16},
\texttt{failed\_count=0}, \texttt{spotlight\_count=16}. Templates are selected
per rhetorical role: \texttt{hook\_title\_intro} for the opening shot,
\texttt{definition\_card} for the Bernoulli explanation,
\texttt{key\_takeaway} for the materials act, \texttt{analogy\_card} for the
winglet--rowing-blade comparison, and \texttt{historical\_moment} for the
aviation-history shot. Critically, \texttt{verified\_count=0}: slot
verification was requested but produced no verdicts, and every
\texttt{verify\_result} is an empty object.

\tracefield{Next decision}
The decoration is structurally complete but semantically unverified. Inspect
and improve the plan before paying for the final render.

\tracestep{5}{Encounter a quality-warning state}

\tracefield{Thinking}
The zero verification count from the previous return is the loose thread in
this run. I have 16 decorated shots that nothing has actually judged, and the
render stage will happily bake whatever text is in those slots into the frames.
Rendering first and inspecting afterwards would be the expensive ordering, so I
inspect now, over all 30 shots rather than only the decorated ones, since a
visual mismatch on a plain shot is just as damaging in the final cut. I scope
the repair to the two dimensions that decoration can actually move ---
decoration quality and visual match --- and request minimal repair with an
immediate re-judge, so that whatever the tool changes is validated in the same
call instead of drifting unchecked into the render.

\tracefield{Observation}
The decorated plan carries no verification verdicts.

\tracefield{Call}
\begin{lstlisting}[style=toolcall]
Tool_Stage4DimensionalRepair(
    failed_shot_ids=["1", "2", ..., "30"],
    target_dimensions=["decoration_quality", "visual_match"],
    minimal_repair=True,
    rejudge_after_repair=True,
    repair_round=0
)
\end{lstlisting}

\tracefield{Return}
\texttt{status=completed\_with\_warnings} after 189.9 seconds:
\texttt{passed\_count=0}, \texttt{failed\_count=30}. Every shot remains below
the strict quality threshold.

\tracefield{Next decision}
A blanket failure over all 30 shots indicates missing evidence rather than
providing a per-shot defect report. Produce explicit self-reflection findings
before attempting another patch.

\tracestep{6}{Hit a typed precondition error}

\tracefield{Thinking}
I want to patch the plan against the reflection findings, and reward shaping is
the augmentation I would like applied so that both passing and failing shots
receive graded per-dimension scores rather than a binary verdict. What I am
about to do, however, is issue synthetic issue identifiers of my own invention.
This is worth recording as an error rather than hiding, because it is the exact
mistake the runtime's typed preconditions exist to catch: I am attempting to
repair against evidence I have not yet collected.

\tracefield{Observation}
No \texttt{self\_reflect\_issues} object exists in the state.

\tracefield{Call}
\begin{lstlisting}[style=toolcall]
Tool_SelfReflect_Patch(
    augmentation_strategy={
        "id": "aug09_reward_shaping",
        "name": "Reward Shaping",
        "focus": "Emit finer-grained reward signal: score every dimension for
                  both passing and failing shots, and state the preference
                  basis used for SFT/RL."
    },
    issue_ids=["sr_001", "sr_002", "sr_003", "sr_004", "sr_005"],
    target_shot_ids=["1", "2", "3", "4", "5"],
    minimal_repair=True,
    requires_downstream_propagation_hint=True
)
\end{lstlisting}

\tracefield{Return}
\texttt{status=blocked\_precondition},
\texttt{missing=["self\_reflect\_issues"]}. The runtime rejects the call with
``insufficient preconditions for Tool\_SelfReflect\_Patch''.

\tracefield{Next decision}
Treat the rejection as an instruction rather than a failure: run reflection
first and patch against the identifiers it actually returns.

\tracestep{7}{Generate the missing self-reflection evidence}

\tracefield{Thinking}
The precondition told me exactly which object is missing, so this turn is fully
determined --- the only real choice left is what to inspect for. I widen the
focus beyond decoration to the five dimensions where an error would be
unrecoverable after rendering: script coverage, shot rhythm, visual-query
alignment, spotlight ratio, and duplication. I deliberately switch the
augmentation strategy from reward shaping to strict semantic coverage. Reward
shaping produces a smoother signal, but at this moment I do not need a smoother
signal. I need a judge that would rather fail a shot than let a semantic
omission through, because a missed narration unit cannot be repaired once the
timeline is composed.

\tracefield{Observation}
The patch tool is blocked, and no reflection object exists.

\tracefield{Call}
\begin{lstlisting}[style=toolcall]
Tool_SelfReflect(
    scope="full_repair_with_downstream_propagation",
    num_segments=6,
    num_shots=30,
    focus_dimensions=["script_coverage", "shot_rhythm",
                      "visual_query_alignment", "spotlight_ratio",
                      "duplication"],
    mode="gemini_augmentation",
    augmentation_strategy={
        "id": "aug01_semantic_coverage_strict",
        "name": "Semantic Coverage Strict",
        "focus": "Re-audit whether the narration covers the core information
                  of the source; prefer failing a shot over letting a semantic
                  omission pass."
    }
)
\end{lstlisting}

\tracefield{Return}
\texttt{status=needs\_fix}, \texttt{passed=false},
\texttt{overall\_score}~$=0.70$. The judge emits 13 concrete issues: 11
\texttt{decoration\_quality} findings (shots 5, 6, 7, 10, 11, 14, 16, 21, 23,
26, 28) and 2 \texttt{visual\_match} findings (shots 14 and 28), all at medium
severity, with \texttt{suggested\_fix="redecorate"}. The summary reads
``13 decoration-quality problems found'', and the state now records
\texttt{repair\_history=[\{round: 0, passed: false, issue\_count: 13\}]}.

\tracefield{Next decision}
Patch exactly these 13 issues. The blanket 30-shot failure of step~5 has been
resolved into a bounded, addressable defect list.

\tracestep{8}{Patch the plan against the returned issue identifiers}

\tracefield{Thinking}
Now I can issue the call I attempted two turns ago, but with the identifiers
the judge actually produced instead of ones I made up. I keep the strict
coverage strategy so that the patch is evaluated by the same standard that
produced the findings. Switching strategies mid-repair would mean patching
against one rubric and being graded by another. I set minimal repair, because
the failures are concentrated in decoration slots and visual queries, and any
edit that ripples into narration text would invalidate the audio I already
synthesized in step~3 at a cost of 14 minutes. I request downstream propagation
hints so the runtime marks which artifacts the patch has invalidated rather
than leaving me to infer it.

\tracefield{Observation}
Thirteen typed issues exist over 13 distinct shots.

\tracefield{Call}
\begin{lstlisting}[style=toolcall]
Tool_SelfReflect_Patch(
    issue_ids=["qj_5_decoration_quality", "qj_6_decoration_quality",
               "qj_7_decoration_quality", "qj_10_decoration_quality",
               "qj_11_decoration_quality", "qj_14_visual_match",
               "qj_14_decoration_quality", "qj_16_decoration_quality",
               "qj_21_decoration_quality", "qj_23_decoration_quality",
               "qj_26_decoration_quality", "qj_28_visual_match",
               "qj_28_decoration_quality"],
    target_shot_ids=["1","2","5","6","7","10","11",
                     "14","16","21","23","26","28"],
    minimal_repair=True,
    requires_downstream_propagation_hint=True,
    augmentation_strategy={"id": "aug01_semantic_coverage_strict", ...}
)
\end{lstlisting}

\tracefield{Return}
\texttt{status=patched}, \texttt{repair\_applied=true}. The patch is indeed
localized: visual queries are sharpened (shot~3's query gains the explicit
``conveying precision and complexity'' qualifier), and shot~2's duration
is re-estimated from 61.11\,s to 60.70\,s. Narration scripts, segment
boundaries, and spotlight assignments are left untouched, so the synthesized
audio remains valid.

\tracefield{Next decision}
Re-run decoration so the repaired shots pick up new templates and slots.

\tracestep{9}{Re-decorate under the repaired plan}

\tracefield{Thinking}
The patch changed the plan, not the rendered overlays, so the decorations
currently in state were selected against text that no longer exists. I re-run
decoration over the full spotlight set rather than only the 13 patched shots:
template selection is a global assignment, and if a patched shot now claims a
template that an unpatched neighbour is already using, restricting the call to
the patched subset would leave that collision in place. The cost is the same
either way, since decoration is cheap relative to rendering.

\tracefield{Observation}
The plan has been patched but the spotlight decorations still reflect the
pre-patch text.

\tracefield{Call}
\begin{lstlisting}[style=toolcall]
Tool_BatchDecorateSpotlights(
    shot_ids=["1","2","5","6","7","10","11","13",
              "14","16","19","21","22","23","26","28"],
    template_policy={
        "select_template": True,
        "fill_slots": True,
        "verify_slots": True,
        "spotlight_only": True
    }
)
\end{lstlisting}

\tracefield{Return}
\texttt{status=success}, \texttt{decorated\_count=16},
\texttt{failed\_count=0}. The re-decoration is visibly better targeted: shot~2's
definition now states both surfaces (``upper surface faster and lower
pressure, lower surface slower and higher pressure'') instead of only the upper
one. Shot~5's takeaways are compressed from full clauses into three scannable
phrases. Shot~6 moves from a generic \texttt{overlay} to a
\texttt{concept\_explanation} type, and shot~10 switches to the more specific
\texttt{historical\_moment\_card} template.

\tracefield{Next decision}
Re-judge the repaired, re-decorated state.

\tracestep{10}{Re-judge all shots against the full dimension set}

\tracefield{Thinking}
This is the verification half of the repair loop, and I want it against all
five delivery dimensions rather than the two I repaired, because a decoration
edit can break timing alignment even when it improves decoration quality. I
hold the threshold at the strict 0.75 and focus the judge on decoration slots,
which is where the patch actually landed.

\tracefield{Observation}
Thirteen shots have been patched and all 16 spotlight shots re-decorated.

\tracefield{Call}
\begin{lstlisting}[style=toolcall]
Tool_Stage4DimensionJudge(
    shot_ids=["1", "2", ..., "30"],
    dimensions=["visual_match", "audio_sync", "decoration_quality",
                "timing_alignment", "render_integrity"],
    threshold=0.75,
    focus_decoration_slots=True
)
\end{lstlisting}

\tracefield{Return}
\texttt{status=completed\_with\_warnings}, \texttt{passed\_count=0},
\texttt{failed\_count=30}, \texttt{\_elapsed\_sec=0.0}.

\tracefield{Next decision}
The zero elapsed time is the informative part: the judge did not re-evaluate,
it replayed a cached verdict. Two of the five requested dimensions
(\texttt{render\_integrity} and \texttt{timing\_alignment}) cannot be scored
before a render exists, so this verdict is structurally unable to improve at
this stage.

\tracestep{11}{Probe whether the failure is real or structural}

\tracefield{Thinking}
Before I spend the rest of the repair budget, I need to distinguish a genuine
quality failure from a stage-ordering artifact. I re-issue the step~5 repair
verbatim. If the state has actually changed, the tool should now do work and
take time. If it returns instantly with the same counts, the verdict is not
tracking my edits and further repair at this stage is pure waste.

\tracefield{Observation}
The re-judge reports the same blanket failure it reported before any repair.

\tracefield{Call}
\begin{lstlisting}[style=toolcall]
Tool_Stage4DimensionalRepair(
    failed_shot_ids=["1", "2", ..., "30"],
    target_dimensions=["decoration_quality", "visual_match"],
    minimal_repair=True,
    rejudge_after_repair=True,
    repair_round=0
)
\end{lstlisting}

\tracefield{Return}
\texttt{status=completed\_with\_warnings}, \texttt{passed\_count=0},
\texttt{failed\_count=30}, \texttt{\_elapsed\_sec=0.0} --- identical counts,
identical zero cost. The hypothesis is confirmed: the warning is structural.

\tracefield{Next decision}
Make one final, explicitly budgeted repair attempt so that the runtime, not the
policy, decides when to stop.

\tracestep{12}{Exhaust the repair budget deliberately}

\tracefield{Thinking}
I could loop here indefinitely, and that is precisely the failure mode I must
avoid: a repair loop that never converges consumes the entire run and delivers
nothing. So I escalate rather than repeat --- I declare this
\texttt{repair\_round=2}, the last one the runtime permits, and widen the
target to all five dimensions so that if any residual defect is genuinely
fixable at plan level, this call has the broadest possible chance to fix it.
Setting the round explicitly hands the stopping decision to the runtime's
budget accounting instead of to my own judgment, which is the safer place for
it.

\tracefield{Observation}
Two repair calls have produced no state change at zero cost.

\tracefield{Call}
\begin{lstlisting}[style=toolcall]
Tool_Stage4DimensionalRepair(
    failed_shot_ids=["1", "2", ..., "30"],
    target_dimensions=["visual_match", "audio_sync", "decoration_quality",
                       "timing_alignment", "render_integrity"],
    minimal_repair=True,
    rejudge_after_repair=True,
    repair_round=2
)
\end{lstlisting}

\tracefield{Return}
\texttt{status=budget\_exhausted}: ``the repair limit of 2 has been
reached; remaining warnings will not be repaired and current quality is treated
as passing (rejection 1 of 2)''.

\tracefield{Next decision}
The quality gate is closed by budget rather than by score. Stop repairing and
move to delivery.

\tracestep{13}{Attempt delivery and learn the real blocker}

\tracefield{Thinking}
With the gate closed I go for the final artifact. I set
\texttt{final\_gate\_required} so the runtime enforces its own delivery
contract rather than trusting my belief that 30 shots are approved, and I
pre-declare the audio mix --- background music at half volume, ducked under the
voice track --- because the narration is the payload and music that competes
with it would ruin an otherwise correct cut. I expect this call to succeed, and
if it does not, its rejection will tell me exactly what the state is still
missing.

\tracefield{Observation}
All 30 shots are treated as approved.

\tracefield{Call}
\begin{lstlisting}[style=toolcall]
Tool_StitchVideo(
    shot_count=30,
    approved_count=30,
    final_gate_required=True,
    bgm_volume=0.5,
    duck_under_voice=True
)
\end{lstlisting}

\tracefield{Return}
\texttt{status=blocked\_precondition},
\texttt{missing=["timeline\_render", "bgm", "rendered\_path\_ready:0/30"]}.

\tracefield{Next decision}
This is the decisive observation of the run, and it retroactively explains
steps~10--12: \texttt{rendered\_path\_ready:0/30} means no shot has ever been
rendered, which is why \texttt{render\_integrity} could never pass and why the
repair loop kept failing on a defect that plan-level edits could not reach. The
correct action was never more repair --- it was to satisfy the two unmet
prerequisites, timeline rendering and background music, in that order.

\tracestep{14}{Satisfy the first prerequisite: compose and render timelines}

\tracefield{Thinking}
This is the expensive, irreversible turn, so I want the runtime to refuse it
rather than produce a subtly broken render. I require both that every shot have
its retrieved visual and synthesized audio present, and that every spotlight
shot carry a decoration --- the second condition is what makes the re-decoration
work of step~9 actually reach the frames instead of sitting unused in state. I
compose all 30 in one batch so that timing offsets are computed against a
single consistent timeline rather than accumulated shot by shot.

\tracefield{Observation}
Assets, audio, and decorations exist in state, but no rendered paths do.

\tracefield{Call}
\begin{lstlisting}[style=toolcall]
Tool_BatchTimelineCompose(
    shot_ids=["1", "2", ..., "30"],
    require_tts_visual_audio=True,
    require_spotlight_decoration=True
)
\end{lstlisting}

\tracefield{Return}
\texttt{status=success} after 378.6 seconds: \texttt{composed=30},
\texttt{render\_success\_count=30}, \texttt{failed\_count=0},
\texttt{concat\_count=16}. The renderer routes by decoration need --- the 16
spotlight shots go through the \hyperframes backend for template compositing,
the remaining 14 through the direct \texttt{ffmpeg} path.

\tracefield{Next decision}
One prerequisite remains: background music.

\tracestep{15}{Satisfy the second prerequisite: retrieve background music}

\tracefield{Thinking}
The music style I requested back in step~1 is still in state, which is why this
turn is a lookup rather than a re-derivation. I query with the full style
description instead of a keyword, since the retrieval index is caption-based
and the tonal arc --- slow building to fast --- is what distinguishes a usable
track from a merely orchestral one. I request more duration than the video
needs so the mixer has slack for crossfades and never runs out of music before
the last shot ends. Setting \texttt{must\_finish\_before\_stitch} makes the
ordering constraint explicit rather than implicit in my turn sequence.

\tracefield{Observation}
Thirty rendered shots exist, but the \texttt{bgm} slot is still empty.

\tracefield{Call}
\begin{lstlisting}[style=toolcall]
Tool_MusicSearch(
    query="epic orchestral, piano lead, strings underneath, tempo from slow
           to fast, grand and warm",
    music_style="epic orchestral, piano lead, strings underneath, tempo from
                 slow to fast, grand and warm",
    target_duration_sec=713.71,
    must_finish_before_stitch=True
)
\end{lstlisting}

\tracefield{Return}
\texttt{status=success}, \texttt{selected\_count=3}. The runtime returns a
sequenced playlist rather than three independent tracks, each with a placed
offset: a 60.0\,s cinematic orchestral opener at $[0,\,60.0]$, a 99.9\,s lyrical
orchestral piece at $[60.0,\,159.9]$, and a 326.6\,s track whose caption reads
``tempo moving from slow to urgent, emotion building'' at
$[159.9,\,486.5]$. Total coverage of 486.5\,s comfortably exceeds the cut.

\tracefield{Next decision}
Both prerequisites are now satisfied. Retry the final gate.

\tracestep{16}{Complete the autonomous final stitch}

\tracefield{Thinking}
I re-issue the exact call the runtime rejected in step~13, unchanged. That it
is unchanged is the point: the call was never wrong, the state was incomplete,
and the two intervening turns closed precisely the gap the rejection named. I
keep \texttt{final\_gate\_required} set so that delivery is validated rather
than asserted, and I will only consider this run finished if the return
contains a real file path and a ready count of 30.

\tracefield{Observation}
The state now contains 30 rendered shots with timeline paths, narration audio,
spotlight decorations, and a three-track music playlist.

\tracefield{Call}
\begin{lstlisting}[style=toolcall]
Tool_StitchVideo(
    shot_count=30,
    approved_count=30,
    final_gate_required=True,
    bgm_volume=0.5,
    duck_under_voice=True
)
\end{lstlisting}

\tracefield{Return}
\texttt{status=success} after 50.1 seconds. The manifest reports
\texttt{ready\_shot\_count=30}, \texttt{missing\_shots=[]},
\texttt{validation\_errors=[]}, and \texttt{duration\_ok=true} on every shot,
with 394.98\,s of stitched video carrying 390.48\,s of narration.
\texttt{concat\_status=success} and \texttt{bgm\_status=bgm\_mixed} across 3
tracks with \texttt{ducking\_fallback\_used=false}, yielding a final
395.0-second delivery:

\begin{center}
\path{final_video_with_bgm_20260824_155554.mp4}
\end{center}

\medskip
\textbf{Takeaway.}
This trajectory exhibits the target inference pattern end to end. The policy
constructs a long-horizon production plan before acting, executes real tools,
and treats every runtime failure as a structured observation rather than a
terminal condition. The invented issue identifiers of step~6 are corrected by
running reflection first. The blanket quality failure of step~5 is resolved into
13 addressable defects by step~7. The non-converging repair loop of
steps~10--12 is diagnosed only when the delivery gate at step~13 names
\texttt{rendered\_path\_ready:0/30} as the true blocker. The policy then
satisfies both missing prerequisites and terminates only after the delivered
artifact exists on disk. Notably, the run consumes its full repair budget
without ever entering an unbounded loop, and the final recovery suffix is a
three-call sequence that repeats no earlier work:

\begin{lstlisting}[style=toolcall]
Tool_BatchTimelineCompose
    -> Tool_MusicSearch
    -> Tool_StitchVideo   (identical arguments to the rejected step 13)
\end{lstlisting}

\end{promptbox}

\section{Visual Inputs to the Evaluation Harness}
\label{app:visual-inputs}

\begin{figure}[htbp]
    \centering
    \includegraphics[width=\linewidth,height=0.88\textheight,keepaspectratio]{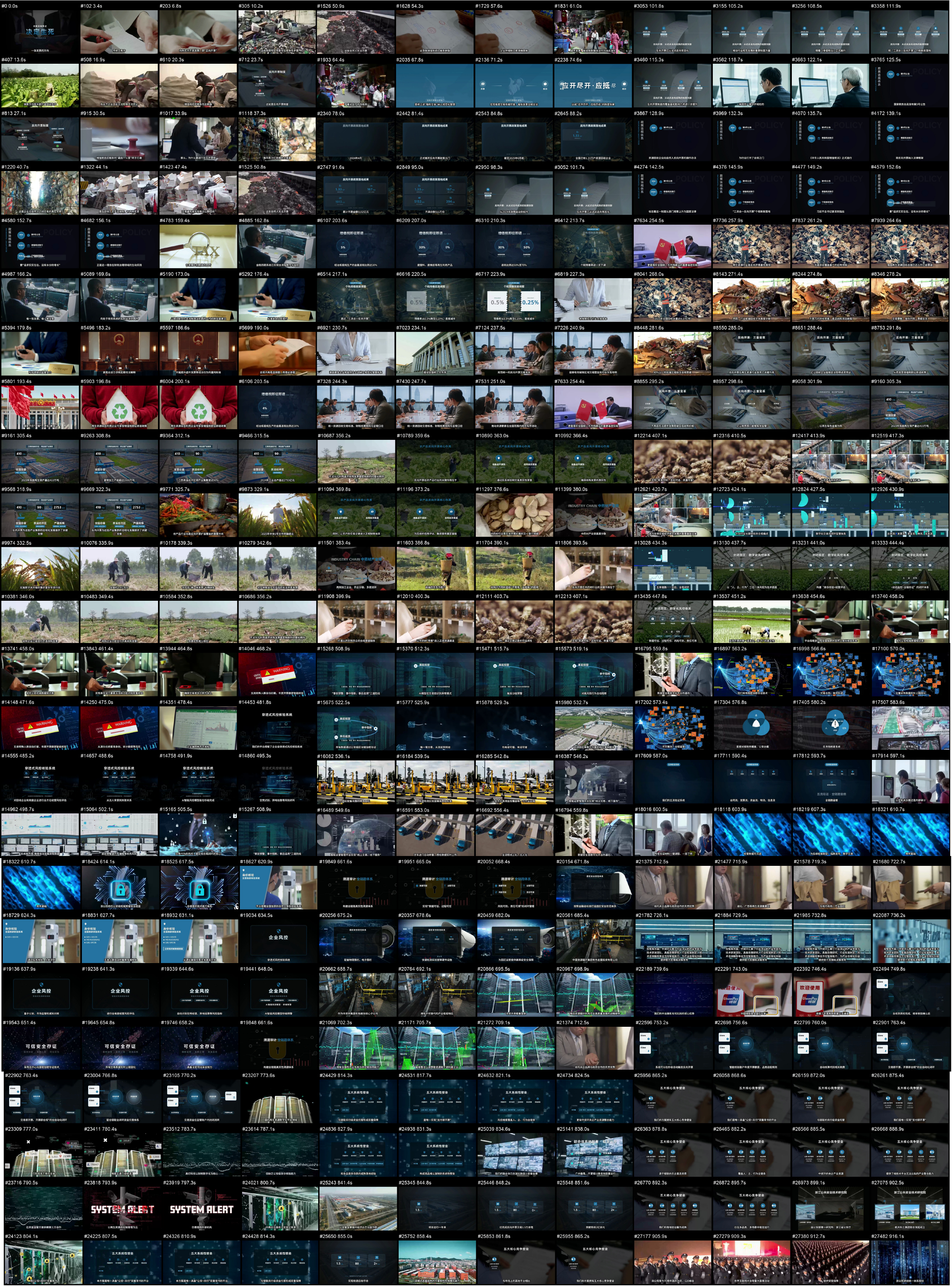}
    \caption{\textbf{Visual Input to the Evaluation Harness: Case 1.}
    A contact sheet of temporally sampled and timestamped video frames used
    for evidence-grounded diagnostic evaluation.}
    \label{fig:harness-visual-evidence-case1}
\end{figure}

\begin{figure}[htbp]
    \centering
    \includegraphics[width=\linewidth,height=0.88\textheight,keepaspectratio]{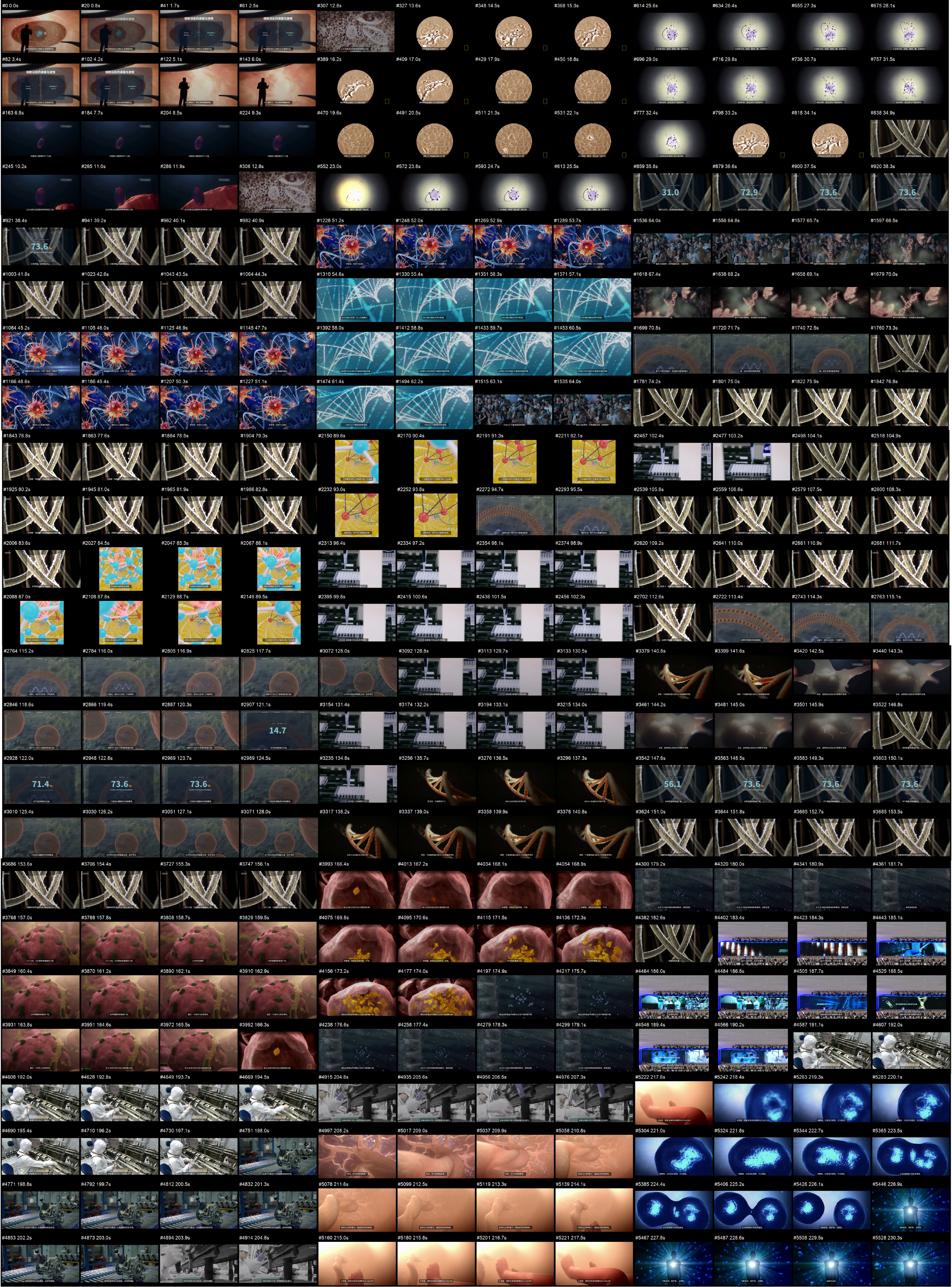}
    \caption{\textbf{Visual Input to the Evaluation Harness: Case 2.}}
    \label{fig:harness-visual-evidence-case2}
\end{figure}

Figures~\ref{fig:harness-visual-evidence-case1} and
\ref{fig:harness-visual-evidence-case2} show two representative
contact-sheet inputs provided to the evaluation harness. Each contact sheet
contains temporally sampled video frames annotated with source-frame indices
and timestamps, allowing the judges to inspect local visual consistency,
timeline progression, long-range coherence, and rendering quality. These
inputs are exposed through read-only, network-free tools and archived together
with their metadata, ensuring that metric decisions are grounded in traceable
visual evidence rather than evaluator-side assumptions.

\section{Limitations}

\agent is limited by the current \hyperframes template library, so uncommon
visual languages may be approximated using the closest available structure,
motion, and style. Expanding the library through automatic template generation could
broaden the design space. The dataset and project-disjoint split cannot cover
all domains, editing conventions, runtimes, or asset distributions, while
verified replay guarantees executability only within the registered runtime.
The quality of long-form videos also depends on the stability of the video
retriever, particularly for instruction-based retrieval. Integrating a trained
video generation model could reduce this dependency and improve robustness.
Finally, VNSM relies on a VLM. Although evidence isolation, frozen anchors,
agreement reporting, and human calibration improve auditability, they cannot
eliminate judge bias or service drift, motivating broader blinded studies and
cross-family audits.

\end{document}